# Improving deep learning with prior knowledge and cognitive models: A survey on enhancing explainability, adversarial robustness and zero-shot learning

Fuseini Mumuni[1]* and Alhassan Mumuni[2]

**Abstract**—We review current and emerging knowledge-informed and brain-inspired cognitive systems for realizing adversarial defenses, eXplainable Artificial Intelligence (XAI), and zero-shot or few-shot learning. Data-driven deep learning models have achieved remarkable performance and demonstrated capabilities surpassing human experts in many applications. Yet, their inability to exploit domain knowledge leads to serious performance limitations in practical applications. In particular, deep learning systems are exposed to adversarial attacks, which can trick them into making glaringly incorrect decisions. Moreover, complex data-driven models typically lack interpretability or explainability, i.e., their decisions cannot be understood by human subjects. Furthermore, models are usually trained on standard datasets with a closed-world assumption. Hence, they struggle to generalize to unseen cases during inference in practical open-world environments, thus, raising the zero- or few-shot generalization problem. Although many conventional solutions exist, explicit domain knowledge, brain-inspired neural network and cognitive architectures offer powerful new dimensions towards alleviating these problems. Prior knowledge is represented in appropriate forms and incorporated in deep learning frameworks to improve performance. Brain-inspired cognition methods use computational models that mimic the human mind to enhance intelligent behavior in artificial agents and autonomous robots. Ultimately, these models achieve better explainability, higher adversarial robustness and data-efficient learning, and can, in turn, provide insights for cognitive science and neuroscience – that is, to deepen human understanding on how the brain works in general, and how it handles these problems.

**Index Terms**—Domain knowledge, cognitive architecture, brain-inspired neural network, explainable AI, adversarial attack, zero-shot generalization.

## 1 INTRODUCTION

### 1.1 State of deep learning

Machine learning is an artificial intelligence (AI) concept that enables computing systems to learn useful relationships from data and then use this information to identify learned patterns and make predictions on new inputs. Deep learning (DL) is a machine learning method that uses a multi-layered arrangement of computational units to learn relevant patterns from data. Many deep learning constructs have been proposed for various use-cases. Some of the most popular ones are deep convolutional neural networks (DCNN) [1], [2] recurrent neural networks (RNN) [3], generative adversarial networks (GANs) [4], deep reinforcement learning [5] and vision transformers [6], [7]). DCNNs are a class of deep neural networks (DNNs) specially designed to handle image data.

Recently, deep learning methods have accomplished remarkable milestones in many critical applications. In medical domains, some state-of-the-art models (e.g., as reported by several researchers [8], [9], [10]) have exceeded human performance in important tasks like clinical diagnosis. Deep learning has also been shown to be capable of outperforming human experts in product design and manufacturing. For instance, Mirhoseini et al. [11] show how DL improves accuracy and efficiency over human chip designers. In arts, deep learning-based AI models have demonstrated performance on par with humans [12]) in creative writing tasks like poetry composition. Also, AI manipulated video and audio content, known as Deepfakes, can look so authentic as to trick humans (see [13]) into thinking that they are real.

### 1.2 Need for knowledge priors and cognitive insights in deep learning

Presently, deep learning methods achieve superior performances than other machine learning approaches. To achieve this, they heavily rely on very large volumes of training data. However, in many practical situations, it is difficult to obtain sufficient training data. Therefore, data insufficiency poses severe limitations to deep learning systems, resulting in significant performance bottlenecks. Inherent noise, irrelevant features and outliers, which inevitably characterize most large-scale training datasets, may also degrade performance. Consequently, knowledge-driven and brain-inspired approaches have been proposed to alleviate the over-reliance on training data.

. [1]*Fuseini Mumuni: University of Mines and Technology, UMaT, Tarkwa, Ghana. E-mail: fmumuni@umat.edu.gh*
*\*Corresponding author, E-mail: fmumuni@umat.edu.gh*
. [2]*Alhassan Mumuni: Cape Coast Technical University, P. O. Box DL 50, Cape Coast, Ghana. E-mail: alhassan.mumuni@cctu.edu.gh;*

In contrast to raw training data that captures no underlying context, prior knowledge and cognitive insights can provide information about real-world relationships or utilize how the mind works to help in connecting the dots between input data and model decisions during inference or training of deep learning models. Besides alleviating the data insufficiency problem — this also helps in mitigating common problems that deep learning systems face in practical applications — adversarial attacks, explainability, generalization to unknown classes and data distributions (i.e., zero-shot learning), and learning with few examples (few-shot learning).

## 1.3 Knowledge representation

In many application domains, prior knowledge is readily available and can be harnessed to augment training data or the training process in deep learning. To achieve this goal, domain knowledge needs to be formalized and structured in a manner that facilitates its integration into deep learning models. Knowledge representation is a principled way of organizing prior knowledge for use by deep learning systems.

**Mathematical equations:** The common ways of representing knowledge include the use of explicit mathematical relations — mostly algebraic or differential equations — to describe real-world systems. These relations are usually employed to constrain the underlying model to conform to some governing mathematical laws. Such an approach can also be used to create surrogate models of complex systems for which data may not be readily available (e.g., [14], [15]). Mathematical knowledge can also be used to perform data argumentation or in simulation environments to create entirely new datasets (e.g., [16] for training machine learning models.

**Knowledge graphs:** Knowledge graphs (KGs) [17], [18] constitute another important class of techniques for knowledge representation in deep learning models. A knowledge graph organizes domain knowledge into structured relationships using nodes to represent physical and abstract entities, interconnecting edges to encode semantic relationships, and labels to define properties of the underlying entities. Besides this graphical form, knowledge graphs can also be represented by textual triples of the form (subject, predicate, object). This representation provides a succinct way of encoding simple relationships, e.g., (*Argentina, WinnerOf, WorldCup*) Ontologies are specialized knowledge graphs which are popular in representing biomedical data. Knowledge graphs have recently achieved promising results in many fields of artificial intelligence. In natural language processing (NLP), they have been successfully applied to improve the performance of recommender systems [19], question answering (QA) [20], [21], language modelling (e.g., [22], and many more.

**Symbolic logic:** Symbolic logic is a popular approach that aims to encode knowledge by using logical representation of simple propositions and their relationships. It exploits symbolic representation of plain sentences to express relationships. Logic representation provides a framework for logical inference that allows logical outcomes to be inferred from given statements using appropriate semantics.

Two forms of logic representations are commonly employed in artificial intelligence and deep learning: propositional and first-order logic.

The incorporation of logical reasoning in deep learning is exemplified by Riegel et al. [23] in the logical neural network (LNN). Each neuron in the LNN is associated with a logical (first-order or propositional) statement, where the network activation functions are usually constrained to perform the specified logical operations on their inputs. Also, since there is a 1-to-1 mapping of neurons to logical units, each connection has a semantic meaning, thus, increasing the interpretability of the network.

**Probabilistic Relationships:** Probabilistic relationships are used to capture uncertainty in knowledge. In fact, knowledge in many real-world settings is characterized by incomplete information, often obtained from partial observation of complex systems. Other systems are affected by factors that cannot be determined a priori, or are liable to changes over time. Machine learning models therefore leverage knowledge represented in probabilistic forms to improve performance in these situations. For instance, in their Probabilistic faster R-CNN, Yi et al. [24] incorporate a probabilistic region proposal network to stochastically predict the objectness of candidate windows in a Faster R-CNN [25] framework used for object detection from remote imagery. The probabilistic model used is Bayesian inference, which assigns confidence scores describing the uncertainty associated with each region proposal, allowing relevant more useful regions to be selected based on their quality, rather than the fixed threshold selection approach that characterizes conventional Faster R-CNN method.

**Implicit knowledge from pre-trained foundation models:** Foundation models are very large-scale neural networks which are pre-trained on vast volumes of multimedia data (e.g., text, image or text-image pairs) readily available in various sources, mostly on the web. They are usually fine-tuned for the intended applications. Owing to their immense size and the large volumes of data they are trained on, foundation models implicitly capture a substantial amount of world knowledge which can be exploited in various downstream tasks like question answering and open-domain object detection. Examples of foundation models include BERT [26], PaLM [27], GPT-3 [28] and InstructGPT [29], the latter two being the key frameworks behind the influential conversational agent ChatGPT. These models excel in zero-shot and few-shot generalization tasks, had have also recently achieved impressive performance on brain stimulus decoding in a zero- and few-shot generalization settings.

## 1.4 Brain-inspired Techniques

Brain-inspired techniques of improving artificial intelligence usually focus on designing systems that leverage the power of the mind to solve complex problems. The two main approaches to realizing this goal are through cognitive architectures and brain-inspired deep neural networks. Brain-inspired cognitive architectures (BICAs) are popular in autonomous robot applications, where the goal is to attain general intelligence. Brain-inspired neural network may, in fact, be part of a BICA framework that would additionally include other subsystems to facilitate tasks like reasoning and planning.

**Brain-inspired cognitive architectures (BICAs):** BICAs describe both the nature of the mind, and equivalent computational models that facilitate the design of artificial intelligence systems that mimic biological cognition. The goal is to achieve in AI systems human-like characteristics such as emotions, memory, reasoning, life-long learning, adaptivity, and general problem-solving ability. Over the years, several BICAs have been proposed. Some of the successful ones in widespread use include Soar [30], Sigma [31], ACT-R [32], [33]. The most important components of a typical BICA include different kinds of working and long-term memory, environment perception and motor or control modules. Figure 1 shows a simplified block diagram of a typical BICA framework. This generic model represents the mind not as a single complex unit, but as a group of distinct parts with specific roles that all work in synergy to achieve desired goals. The most important components of a typical BICA include different kinds of working and long-term memories, environment perception and motor or control modules.

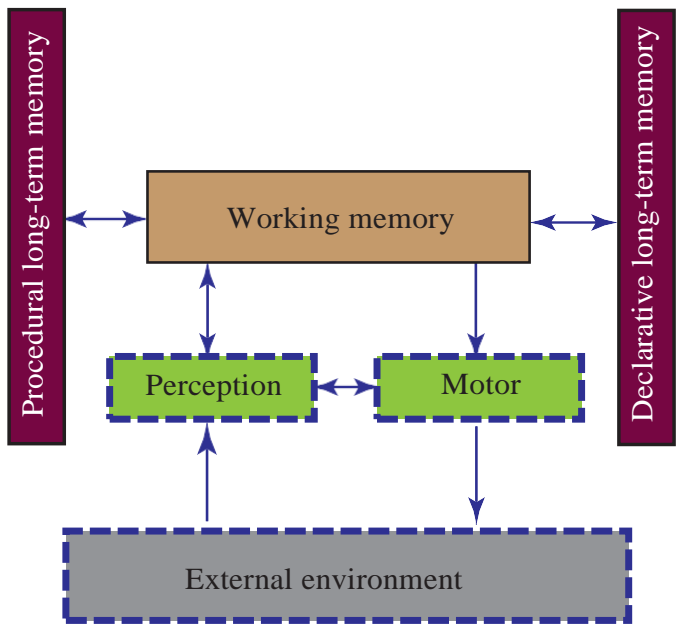

Figure 1. Simplified block diagram showing the main components of a cognitive architecture.

**Brain-inspired deep neural networks:** Despite their initial biological motivation, current state-of-the-art DNNs at best, represent a rather oversimplified abstraction of real neural networks. For instance, the more biologically realistic 'integrate-and-fire' neuron activity is only approximated by spiking neural networks [34]. Still, deep convolutional neural networks (DCNNs) are a class of artificial neural networks that have a fair level of similarity in organization to the human visual cortex. From this perspective, convolution operations may play the role of simple cells while pooling layers perform the action of spatially invariant complex cells [35]. Like their biological counterparts, DCNNs are adept at vision tasks. It is well known that the immense power of the human visual cortex is partly owed to its intricate organization that includes lateral and feedback or recurrent connections. However, traditional DCNNs maintain exclusively feedforward connections. Unsurprisingly, architectural changes to accommodate more cortex-style interconnections in DCNNs have yielded promising results [36], [37], [38], [39], [40].

Other brain-inspired approaches of improving DNNs' adversarial robustness and zero-shot or few-shot generalization are being actively investigated. Some of the key approaches are discussed in Section 4.1.2. They include 1) modification of the backpropagation training scheme into a brain-informed predictive coding scheme; 2) improving the biological plausibility of the neuron model; 3) ensuring biological realism of the overall network architecture.

Despite the numerous studies confirming the immense potential of incorporating brain-inspired enhancements in artificial neural networks, works on this front still lag far behind conventional deep learning methods both in terms of volume and overall accuracy. This situation can be explained by the fact that available knowledge on the human brain is still far from complete. Fortunately, advancements in brain-inspired DNN techniques have now widely been explored in a bid to explain the structure and function of the brain. Using brain-inspired DNNs as surrogates allow non-invasive and easily adaptable simulations to be performed in order for neuroscientists to gain deeper insights into the organization and innerworkings of the brain. Such frameworks may be designed to simulate single biological neurons (e.g., [41], neuronal circuits [42], and whole brain regions (e.g., [43]. The insights learnt from these simulations could in-turn, help to achieve better biological realism in artificial neural networks. In CORnet-S [44] leverage neuroscience and deep learning principles to design a brain-inspired convolutional neural network anatomically aligned with real brain regions and recurrent connectivity. This model achieves high performance on image classification datasets and at the same time provides further insights into the cognitive processes of the brain.

## 1.5 Motivation and Outline of work

When faced with unexpected situations such as adversarial attacks, data-driven deep learning models produce woefully poor results that can potentially lead to dire consequences. Additionally, deep learning systems struggle to cope in situations where certain categories are few or absent in the training data. To handle this problem, machine learning frameworks must be endowed with zero- or few-shot learning capability. Moreover, to ensure understanding and promote trust, especially in human-robot interactions, deep learning systems must be able to explain their decisions to human stakeholders. Based on recent promising research in the areas of knowledge-informed machine learning and brain-inspired cognitive systems, we believe that today's deep learning models can benefit immensely from widely available world knowledge and by incorporating useful concepts from biological systems to better solve the aforementioned problems. Finally, just as knowledge of the brain's structure and mechanisms underlying its function are critical in improving deep learning, it is also vital to be able to use insights from deep learning to improve understanding of cognitive neuroscience and psychology. Fortunately, simulations of brain-inspired deep neural networks are already beginning to illuminate cognitive science, providing vital insights needed to better comprehend and apply new techniques to develop useful interventions in areas like psychology, medicine and biology. Furthermore, pre-trained cross-modal vision-language networks have yielded promising results in brain stimuli decoding tests. Motivated by these prospects, we present the state-of-the-art research work in these directions, highlighting the significance of the results as well as the key innovations required to attain those achievements. Different from all previous surveys, this paper is particularly focused on knowledge- and cognitive-driven approaches that aim to mitigate problems of adversarial attacks, explainability and zero- or few-shot learning in deep learning frameworks. We also discuss approaches that leverage these developments to provide insights for better understanding of cognitive science.

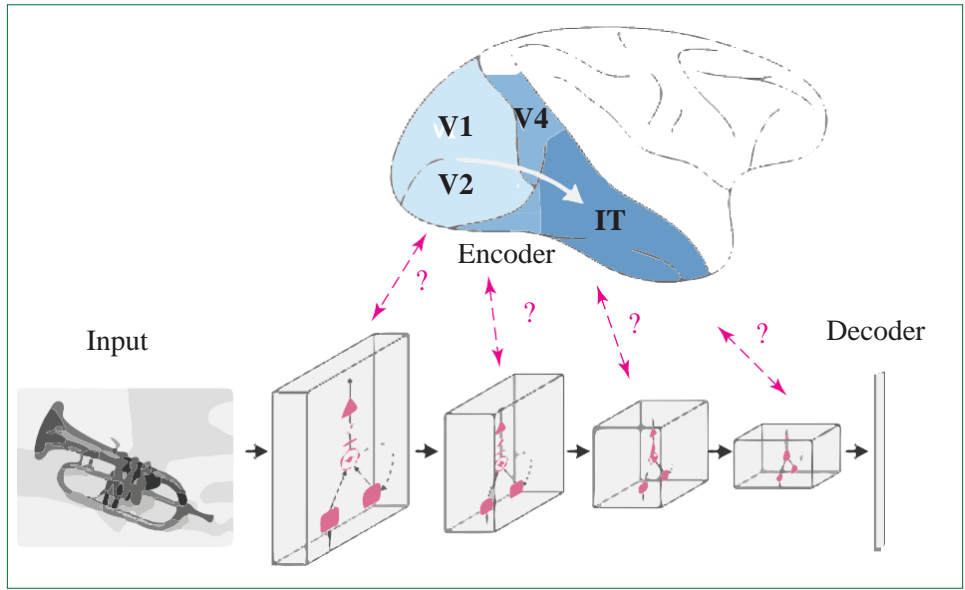

Figure 2. Architectural alignment of CORnet-S [44] with cortical areas V1, V2, V4, and IT in the ventral stream of the brain facilitates brain-like performance and provides further insights on its inner workings.

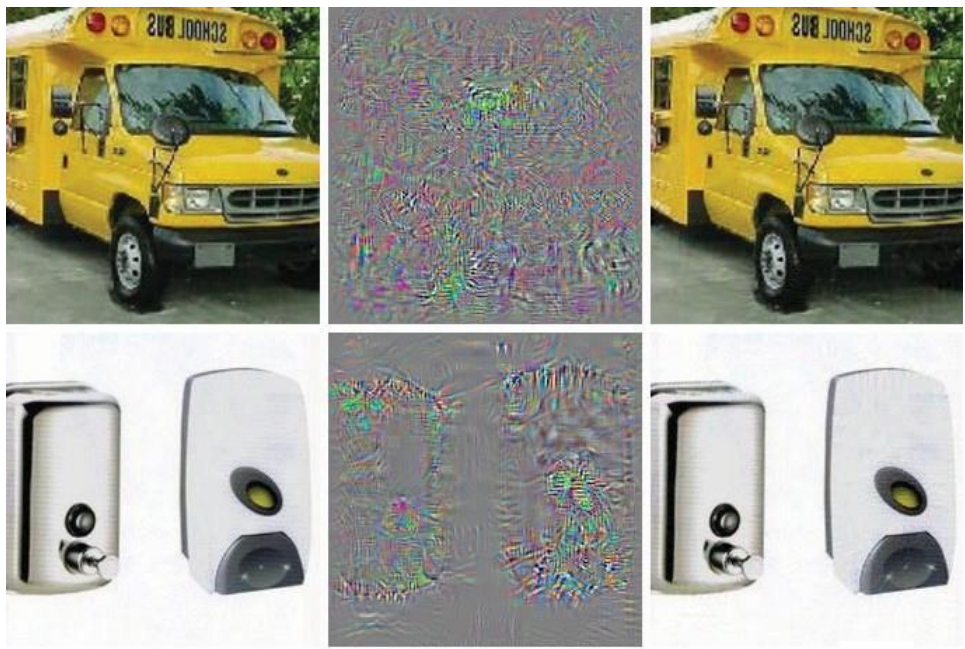

Figure 3. Illustration of adversarial attack: After corrupting the images on the left with adversarial noise (middle) at the pixel level, the resulting images (right) look unchanged from the perspective of a human observer, but the deep learning classifier wrongly classifies both images (i.e., on the right-hand side) as ostrich [45].

The rest of this paper is organized as follows: Section 2 provides the background to the aforementioned issues. Section 3 presents methods of improving adversarial robustness, explainability and zero-shot learning using prior knowledge. In Section 4 we present methods for improving deep learning that rely on insights from cognitive science, and particularly outline the importance of cognitive architectures and brain-inspired DNNs. In addition, we describe some of the important challenges limiting the application of these techniques, and identify ways of overcoming these challenges. We present brain decoding techniques that leverage prior knowledge from pre-trained vision-language models in Section 5. We discuss salient issues and future outlook in Section 6 and conclude in Section 7. The general outline of the paper is shown in Figure 4.

## 2 BACKGROUND

### 2.1 Adversarial Attacks

Deep learning is undoubtedly very powerful in many use-cases. However, achieving human-level versatility and reliability in a broad range of tasks still remains a distant prospect. Deep learning systems can sometimes make bizarre predictions (e.g., [46] when they encounter unknown situations, resulting in catastrophic outcomes. More worryingly, state-of-the-art deep learning models can often be tricked by adversarial attacks, causing them to make utterly wrong predictions by modifying their inputs in subtle and seemingly harmless ways. This weakness can be exploited by human actors with malicious intents to mislead DL systems by corrupting their sensory inputs or by directly interfering with the prediction process itself. Adversarial attacks were first reported by Szegedy et al. [45] in AlexNet on the ImageNet dataset. By adding minor artificial noise to test samples from ImageNet, they observed that the AlexNet classifier would make wildly incorrect predictions, such as misclassifying the truck shown in Figure 3 as an ostrich (and even doing so with a high probability score). Since then the phenomenon has been extensively studied in computer vision (e.g., [47], [48], [49], [50]) and natural language processing (e.g., [51], [52], [53]) domains.

**Conventional defenses and limitations:** Popular defenses against adversarial attacks include 1) adversarial training (e.g., [54], [55]), where adversarial examples are used during training; 2) data augmentation (e.g., [56]), which attempts to create pseudo-adversarial samples for training; and 3) certified defenses (e.g., [57], [58]), which provide guaranteed robustness to bounded adversarial noise. However, despite their popularity, conventional data-driven strategies designed to defend against adversarial attacks lack the reasoning ability to detect illogical predictions, and hence, have limited applicability in most practical use cases. For example, approaches based on adversarial training [59], [60]) are simple to implement but they require large amounts of adversarial examples. Moreover, the training process can induce unintended bias which may harm performance (as per recent studies, e.g., [61]) on clean samples. Furthermore, such defenses are only effective against adversaries that they have been specifically trained to overcome, and even with that they have been shown to be vulnerable to counter-attacks [62], [63]). Conventional data augmentation may experience the robust overfitting [61]) problem, which leads to a trade-off between training and test-time accuracies. In achieving guaranteed robustness, certified defenses sacrifice computational time and standard accuracy on clean inputs, or are restricted to model-specific designs. To alleviate these problems, adversarial defenses can leverage prior knowledge to achieve higher robustness. Techniques for realizing this goal are covered in Section 3.1.

### 2.2 Explainable artificial intelligence

In practical applications, deep learning systems are typically employed to autonomously make decisions and take appropriate actions, or to aid practitioners to arrive at desired decisions. To achieve state-of-the-art performance, most of these approaches use highly sophisticated black-box models whose decisions cannot be understood by human experts. Despite their enormous capabilities and their widespread practical applications in many important domains, the inability to provide explanations on the rationale of their decisions is a major concern for their use in some high-stake applications. As a consequence, practitioners and other stakeholders have questioned the suitability of these systems for safety-critical applications like autonomous driving, medical diagnosis and treatment recommendation, where the stakes may be too high to trust opaque decisions. To meet this need, explainable Artificial Intelligence (XAI) techniques have been proposed to provide human-interpretable explanations of the decisions of deep learning systems. Explainability and interpretability are terms that are often used synonymously. More technically, interpretabilty is associated with models which highlight specific features as contributing to the decision, or whose decisions can readily be understood. On the other hand, explainability applies to models which provide explicit descriptions of the source of their decisions.

Understanding the decision process of DL models creates the trust needed to fully rely on these models to perform critical functions. Even in situations where artificial agents work in tandem with human experts and the models' predictions are only meant to assist humans arrive at final conclusions, it is vital to understand why certain decisions are made, as well as understand the limitations and biases of the machines' decision-making process. This understanding enables humans to be able to put the machines' decisions in context, and also determine which predictions of the machine to accept or reject, and which outputs to probe further. Thus, model explainability or interpretability is essential to collaborative decision-making (see [65], [66]) by experts and intelligent agents. Furthermore, when deep learning models make wrong decisions that lead to adverse outcomes, explainability can help in determining who to apportion blame (see [67], [68]) or settle legal [69], [70]). Hence, the demand for explainability is growing strongly in recent times [71]. As shown in Figure 5, deep learning methods achieve remarkable accuracies but they require extremely large and complex models that rely on large-scale training data for their impressive performances. Owing to their immense complexity, as illustrated by Yang et al. [64] (see Figure 5), these high performing deep learning systems are characteristically opaque. In contrast, simpler and less accurate frameworks like decision tress, linear models, and simple knowledge-based methods like Markov Logic Networks, Bayesian networks and logic rules are, by their nature, more transparent and inherently interpretable. So, ideally, it would be better to enhance high-performing deep learning models with explainability, rather than adopting low-accuracy white-box models.

**Conventional approaches to explainable AI:** Feature attribution methods are among the most popular inter-

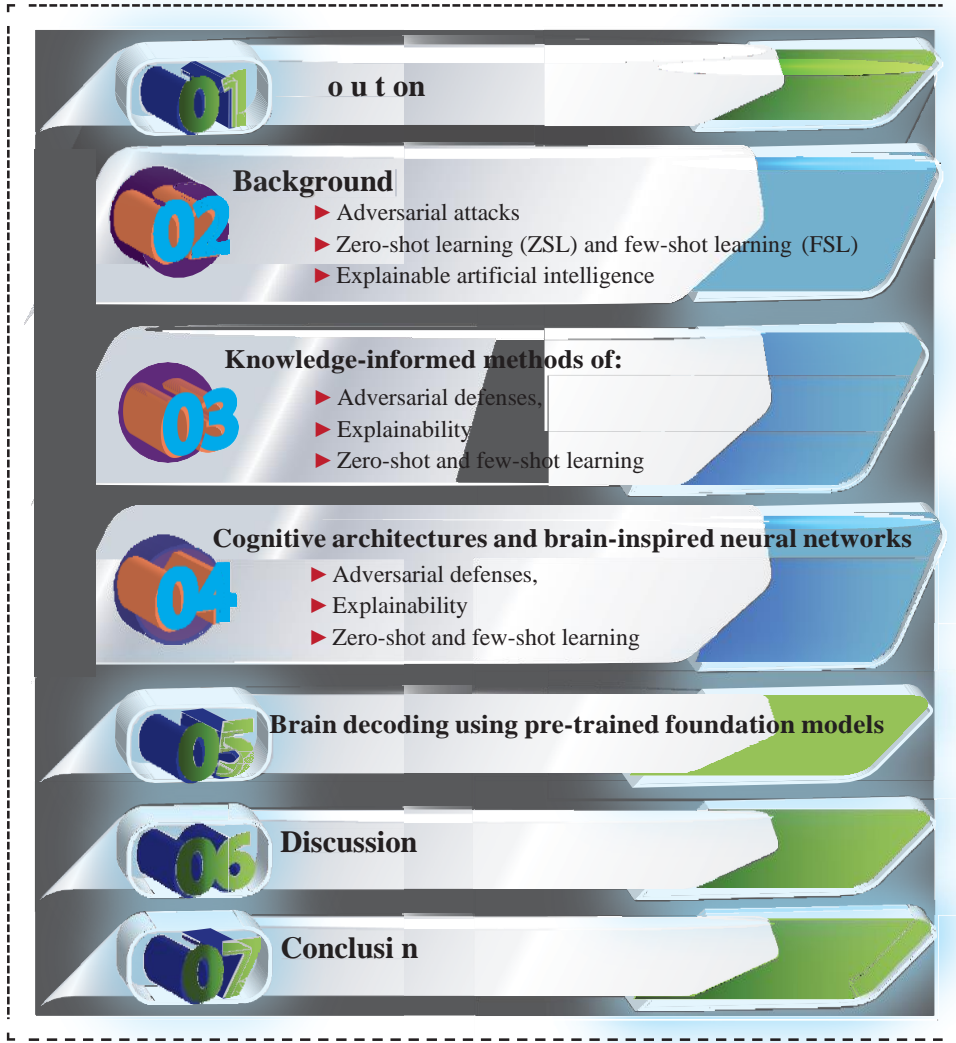

Figure 4. Structure and outline of the paper.

pretability techniques in use. They rely on mapping specific features from the input data to model decisions using a feature-importance score usually computed at the pixel or patch level. Popular among feature attribution techniques include saliency visualization methods based on gradient computations – e.g., Grad-CAM [72]; Integrated Gradients (IG) [73] – and perturbation-based approaches such as locally interpretable model-agnostic explanations (LIME) [74]) and Shapley values (SHAP [75]) which rely on perturbing the input and examining the corresponding effect on model decisions. Attribution methods are utilized in diverse machine learning applications. For instance, in image classification (e.g., [76]), it is used to the highlighted (in the form of heatmaps) regions of the input image that are believed to be responsible for a positive prediction. In natural language processing (NLP) it is widely utilized in tasks like question answering (e.g., [77]) and visual question answering (VQA) (e.g., [72], [78]), the resultant heatmaps usually represent couples of words deemed relevant to a model's answer or image regions that most strongly influence a model's answer. Unfortunately, the quality of the resulting explanations is limited. Specifically, these methods cannot provide the required logic nor the expressivity needed to produce easily-comprehensible explanations from a human perspective. Furthermore, relevant features can be degraded by noise [79], leading to impaired explainability performance. Another major limitation of this category of techniques is that it only highlights regions the model considers important to the decision but the actual reason of the decision is not provided. Thus, attribution methods are unable to provide any meaningful connection between real-world contexts and the corresponding predictions. Multiple studies (see [80], [81], [82]) have also outlined other limitations of these methods. For instance, Figure 6 [83] shows the result of a positive Covid-19 diagnosis from chest X-ray images. The explanation produced by Grad-

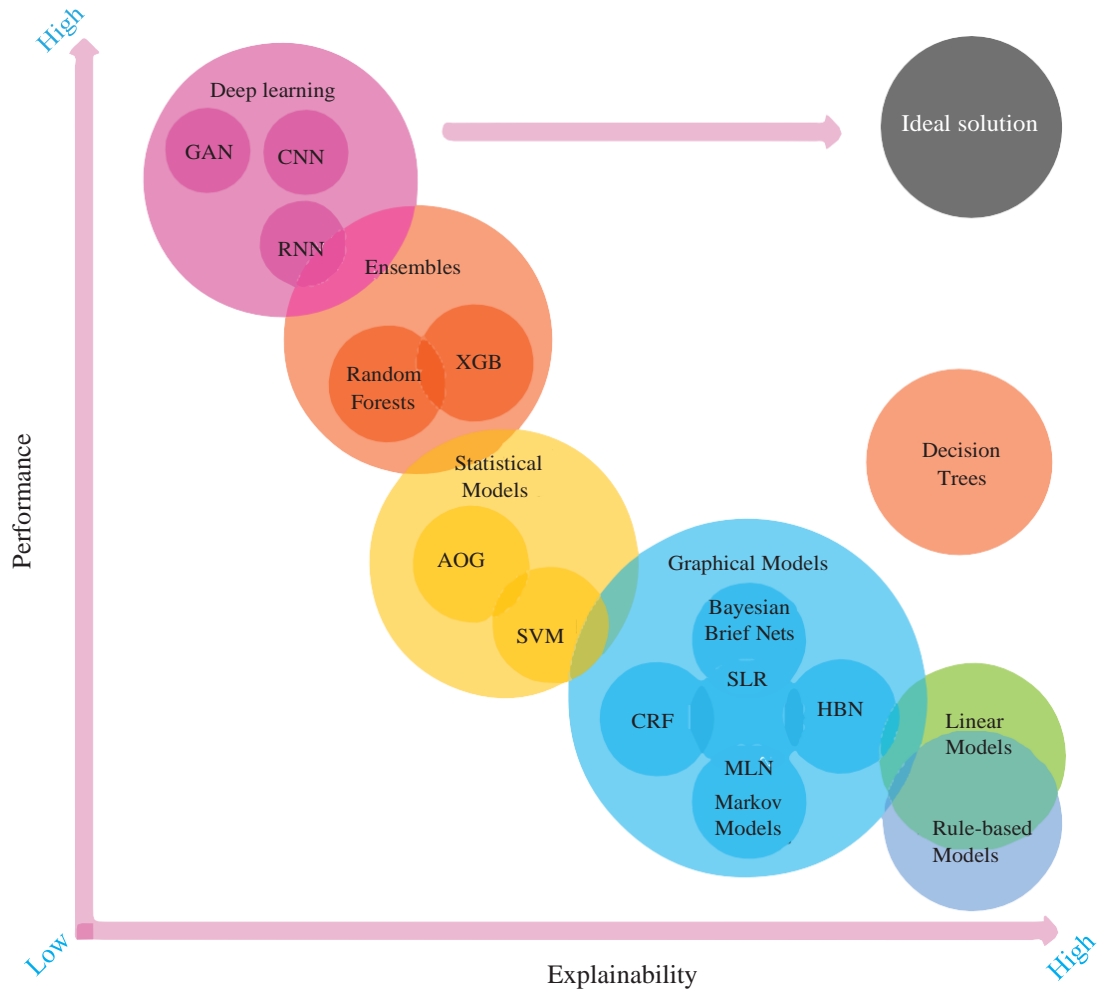

Figure 5. Performance and explainability relationships of machine learning frameworks [64].

CAM highlights features in the image which are clearly unrelated to the disease. These features are in fact textual markings of the letters "PORTABLE UPRIGHT" in the X-ray scans meant to indicate the right-hand side of the patient in the image. As shown by the portions highlighted brown in the top left corner of Figure 6 (b), the model incorrectly assigns importance to the textual inscriptions rather than the actual features that matter to the diagnosis. The model has thus, learned spurious correlation between the text and the disease due to the coincidence of occurrences of this label and positive disease cases in the training data. This flaw, known as the Clever Hans problem [84] is very common in image classification. However, as explained in Section 3.2, knowledge-informed interpretability methods like logical rules or human-in-the-loop techniques overcome these kinds of problems.

Some interpretability methods [85], [86], [87]), called surrogate models, rely on training an inherently interpretable model as a surrogate to provide similar (but explainable) predictions as the original. However, this approach has clear limitations. First, performances of large and complex models designed for high performance cannot be replicated by simple interpretable models (models must be simple enough to facilitate interpretability) since the latter cannot achieve satisfactory performance on complex tasks.

Prototypical Part Networks (ProtoPNets) [88], [89] are another popular group of frameworks that compose explainable decisions from composite parts. They are usually convolutional networks trained end-to-end on images to make predictions and simultaneously learn part-based prototypes associated with the object categories. Similarities with learned prototypes are then used to interpret model decisions at inference time. Given the superior performance of vision transformers in image classification and object detection, recent design [90], [91]) propose a transformer-based prototypical network, or ProtoPFormer, for interpretable image classification. The main problem faced with transformer frameworks is that, due to their tendency to encode long-term dependencies, transformers ultimately learn irrelevant background prototypes. ProtoPFormer introduces a parallel global prototype branch in the network to guide a local branch to focus on the foreground features. Generally, prototype-based explainability methods are very promising as they can identify high-level features that inform model decisions. Furthermore, unlike post hoc methods such as surrogate models, prototypical methods are inherently interpretable and do not rely on external interpreters for explanations.

Inherent limitations of these conventional approaches call for alternative methods and the incorporation of prior knowledge is one of the most promising options available. For instance, surrogate models may not adequately reach the performances of state-of-the-art deep learning models if they maintain acceptable complexities that allow high degree of explainability. Saliency maps may highlight image regions that strongly influences a model's classification decision but such a map could wrongly focus on an irrelevant feature that is usually present around the relevant feature. Methods based on prototypical parts may still require additional logic to associate the relevant information for human understanding. Even the decision logic of inherently explainable methods is only obvious to people with fair knowledge of the models and hence their explanations of such models may not be relevant to end-users whose expertise is in other areas like medicine or law and justice. Ultimately, leveraging domain or expert knowledge to enhance explainability of deep learning models is often necessary to achieve desired results. Consequently, a wide variety of knowledge-driven approaches have been proposed to improve explanations of deep learning models by incorporating prior knowledge to meet human-centric explainability objectives.

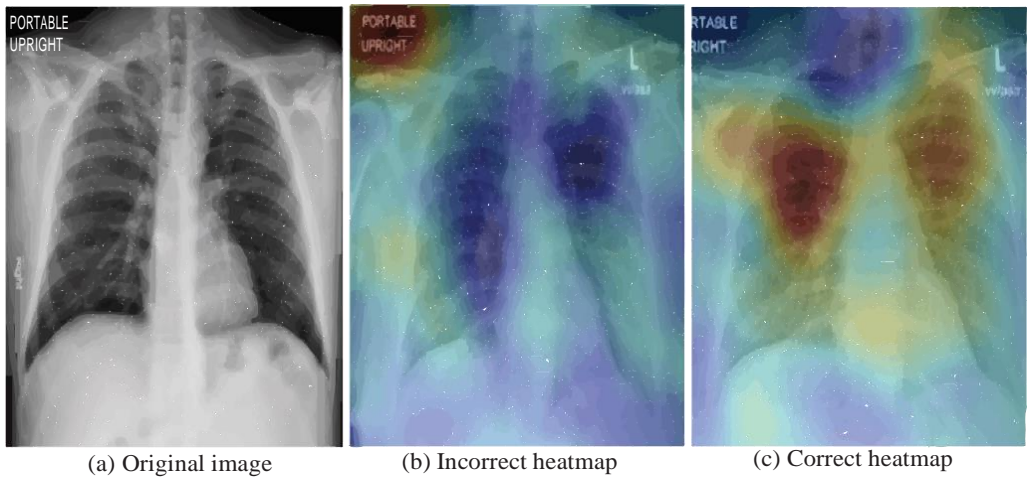

(a) Original image     (b) Incorrect heatmap     (c) Correct heatmap

Figure 6. A chest X-ray image [83] from Covid-19 patient in (a) and a heatmap produced by Grad-CAM [72] incorrectly highlighting the text in the upper left corner as explanation for the positive diagnosis (b); The image on the right shows the correct heatmap capturing the features relevant for the diagnosis.

### 2.3 Zero-shot and few-shot learning

*Zero-shot Learning* (ZSL) is an approach that aims at equipping machine learning models with the ability to generalize from observed classes during training to unseen targets at inference time. Zero-shot Learning can be considered in two ways: (1) The classic concept of ZSL, which assumes different distributions of source and target classes (that is, in deployment, the classifier is required to deal with objects from entirely different categories from those encountered during training); and (2) the more practical setting—called generalized ZSL—which considers samples from both source and target classes at test time (that is, the classifier will handle objects from both seen and unseen classes). Since the datasets used for training deep learning models have a finite number of samples and object classes, deployed systems in the open world are likely to encounter objects belonging to both observed categories and categories not present in the training dataset. This problem is treated as a generalized zero-shot generalization task. In learning under imbalanced or limited data regimes, test time prediction of samples from minority classes is a few-shot generalization problem. Given the reliance on data, it is clear there will be a challenge when a deep learning-based classifier is tasked with recognizing categories for which only a few examples are presented during training.

**Overview of conventional approaches:** Here, conventional approaches are understood as methods that do not leverage world knowledge. Zero-shot learning methods usually rely on common attributes between observed and unseen classes to facilitate generalization from the former to the latter. In image classification, some traditional approaches involve learning the desired compatibility function that maps between visual appearance and semantic attributes. To accomplish this task, these methods either map from visual to semantic (e.g., [92], [93], [94]) or from semantic to visual (e.g., [95], [96]) embedding space. Mapping from visual to semantic space incurs the so-called hubness problem [95]) – which causes the learner to be biased towards a small subset of classes – while the opposite transformation (i.e., semantic to visual) is difficult since one class attribute or description can possibly match the visual appearance of different images. Overall, this line of approaches has a severe performance limitation, especially in the generalized zero-shot learning setting, where there is a significant bias towards observed classes.

Subsequent works [97], [98] have been proposed to address this shortcoming by leveraging generative adversarial networks (GANs) to synthesize samples of unseen categories conditioned on the distribution of class attributes. These methods create synthetic images from scratch through data augmentation techniques. However, the generated images may lack sufficient details to adequately represent the target features, resulting in degraded performance [99] and may suffer from mode collapse [100]. Although other families of approaches exist, prominently those that leverage the complementary benefits of GANs and or variational autoencoders (VAEs) methods (e.g., [101], [102]), the performance difference between generic machine learning and

zero- or few-shot learning is still wide. Moreover, most of these studies have been carried out on standard datasets. Thus, the models would inevitably experience further performance drop when deployed in practical autonomous robotic applications.

## 3 USING KNOWLEDGE TO IMPROVE ADVERSARIAL ROBUSTNESS, EXPLAINABILITY AND ZSL

### 3.1 Using prior knowledge to overcome Adversarial Attacks

Knowledge-based approaches to adversarial defenses can perform consistency checks (e.g., [103]) on models' decisions to ensure conformity with existing domain knowledge. Results that violate logical reasoning or other forms of knowledge are thus corrected or discarded. Some of the most effective approaches are based on object co-occurrence relationships, logic rules, and compositional part-based reasoning.

**Co-occurrence relationships:** Contextual knowledge in the form of category co-occurrence relationships (e.g., [104])—Figure 7 has—been shown to be a highly effective way of detecting and preventing adversarial attacks. This method is based on the premise that certain classes of objects (e.g., an ant and an elephant or a bus and a toothbrush) are highly unlikely to be seen together in a given scene. This conflict may simply be due to mismatch scale or semantic context. With the approach, the decision is probed whenever a co-occurrence constraint is violated. Although counter-attacks against this kind of defense have been proven (e.g., [105]) to be successful, the attackers require the data distribution of the target model to be known. Also, the attacker must know the concept that defines co-occurrence context from the victim's perspective. These requirements are however, difficult to meet in practice. Hence, co-occurrence relationships can offer potent defense in many real-world situations. To further improve robustness, knowledge-informed data augmentation (e.g., [106]) that employ co-occurence logic have been utilize for creating artificial categorical collisions which a network then learns to separate in order to resolve ambiguities.

**Logic rules:** Recently, the use of domain knowledge represented by first order logic rules has shown a significant promise in identifying incoherent, adversarially-influenced predictions in image classification tasks. For instance, Melacci et al. [107] exploit this kind of domain knowledge to enforce logical constraints on training data, encouraging the data to assume the appropriate marginal distribution. During inference, spurious predictions that do not align with training distribution are discarded. Although the method is flexible and can be applied to a wide range of problems, the logic rules needed to capture domain-specific properties must be formulated with expert knowledge. This requirement limits the application of the technique to situations where such expertise is available. To overcome this challenge, Ciravegna et al. [108] propose logic explained networks (LENs) which can directly learn first order logic rules from input data – provided in the form of human-interpretable predicates – to capture object relationships in the target domain. LENs also facilitates interpretability by exploiting first-order logic rules in the output space.

In LOGICDEF (Figure 8), Yang et al. [109] are inspired by the fact that the human visual system easily detects incoherent scenes by mining contextual cues from a powerful scene representation that organizes its elements in a structured hierarchy of objects and their relationships. From a scene graph, LOGICDEF extracts logic rules about objects and their relationships and combines this information with commonsense knowledge derived from ConceptNet [110]—a large commonsense knowledge graph of frequently-used words and phrases collected from diverse sources—to enforce contextual consistency constraints over the scene elements. Besides being able to detect visually implausible predictions, LOGICDEF is also interpretable.

Logical reasoning techniques (e.g., [111], [112]) have also shown promise in facilitating certifiable adversarial defenses. For instance, Zhang et al. [112] utilize a graph convolutional neural network (GCN) for learning semantic features while probabilistic graphical logic framework in the form of Markov logic network (MLN) is used to enhance reasoning over plausible outputs. Similarly, Yang et al. [111] ncorporate MLN to perform reasoning in a deep learning setting. Although still in the early stages, these studies represent an interesting direction of research on certified robustness.

**Compositional part-based reasoning:** Finally, an emerging line of research in adversarial robustness is motivated by the part-based reasoning [113], [114], [115] paradigm employed by biological vision. Human vision is extremely powerful. It has been developed through years of evolutionary adaptation. It is argued that the visual system prefers to recognize objects based on a representation that partitions them into distinguishable parts to allow easy identification using information on the relationships of these components. For this reason, biological vision is extremely robust and is difficult to be fooled by most practical adversarial attacks that modify only part of the target [116]. While there is substantial evidence (e.g., [113], [114], [115]) supporting the aforementioned view that much of the remarkable robustness of the human visual system owes its strength to the compositional or part-based inference paradigm, deep learning methods aiming to circumvent adversarial attacks rarely utilize these robust representations. However, recent studies have demonstrated that part-based reasoning can achieve high generalizability generalization to varied and unknown adversaries (including adaptive attacks), unlike conventional methods which are typically designed to handle specific attacks. In their ROCK framework, Li et al. [117] propose an image classification model based on "Recognizing Object by Components with human prior Knowledge". The method first divides objects into distinct parts, then use this information to recognize whole objects based on part relationships that are predefined through human prior knowledge. ROCK uses a judgement block to produce the final prediction by first computing independent part-based prediction scores and then evaluating the part-linkage according to commonsense knowledge rules. Empirical results presented by the authors show that the approach is effective against different kinds of adversarial attacks and offers significantly better robustness than conventional defenses. Furthermore, Rock does not compromise clean accuracy on benign images.

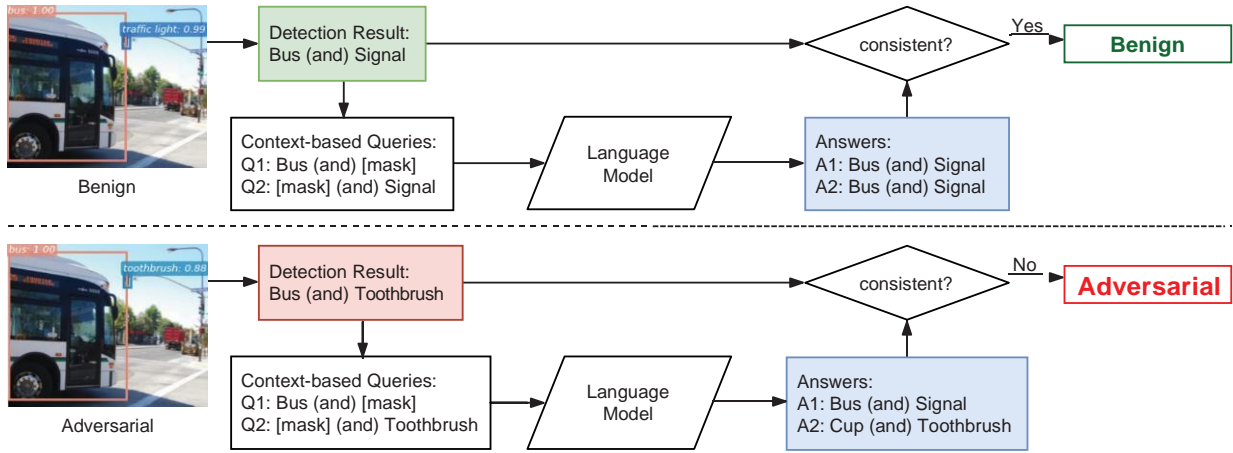

Figure 7. Using co-occurrence logic (e.g., [104]), it is easy to detect the inconsistency of a bus and a toothbrush co-existing in a scene. A more logical combination would be a bus and a traffic sign. The implausible results are discounted as adversarial.

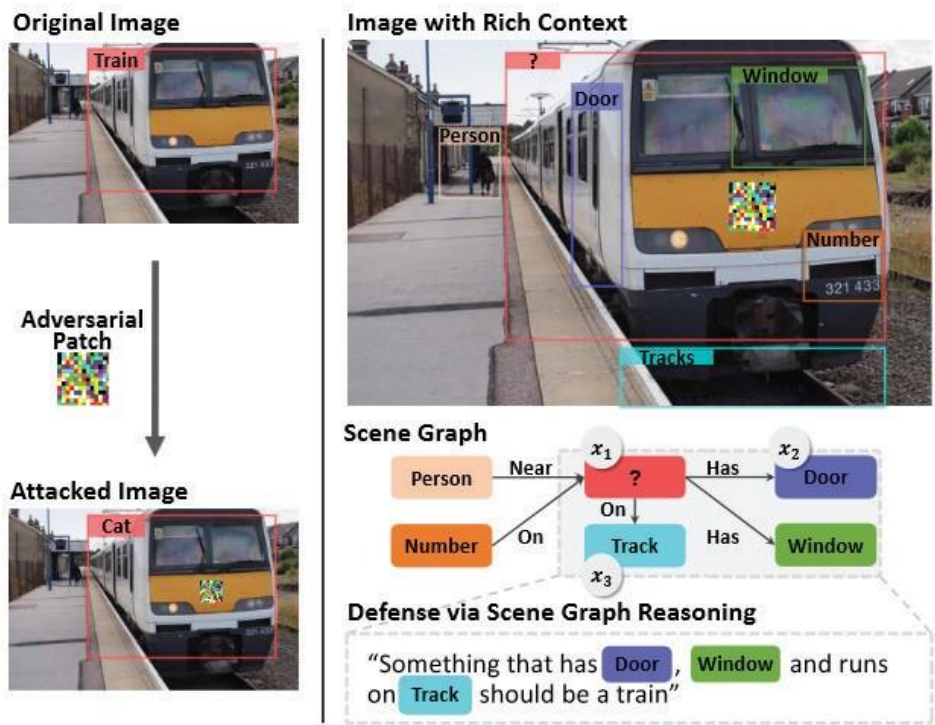

Figure 8. Left: The adversarial patch printed on the train causes a deep learning classifier to misclassify it as a cat. But, using logical context (Right), LOGICDEF [109] correctly discovers that the object must be a train

## 3.2 Knowledge-informed explainability methods

High-level concepts and human-in-the-Loop: Knowledge in the form of human-defined concepts have shown promising results in enhancing explainability in computer vision applications. Concepts (see Concept Bottlenecks [118] and Concept Activation Vectors [119] for typical examples) in this context refer to high-level semantic information such as the stripes of a zebra, the distinctive yellow beak that identifies a parrot, or the narrowing of the space in the knee joint that characterizes arthritis disease. This class of explainability frameworks [118], [120], [121], [122], [123] are popularly referred to as Concept Bottleneck Models (CBMs). CBMs operate by first predicting labels for human-understandable high-level concepts from the input data, and then using the information on predicted concepts to make the final prediction about the category of the image. Usually, CBMs require expert annotation of relevant concepts for training. This makes the approach labor-intensive and expensive to implement.

Human-in-the-hoop explainability methods leverage knowledge obtained directly from a human participant in the training or inference phase of deep learning systems. The premise here is that, the accuracy of a predictions is contingent on the corresponding explanation (i.e., reason for that decision) being correct. Therefore, by observing accompanying explanations, a human participant can intervene to correct explanations that do not match the particular example, causing the prediction itself to change. For instance, by allowing the user to correct mistakes over concept predictions, [118] (see Figure 9) achieves significant accuracy improvements at inference-time. Following a similar line of thought, other works (e.g., [124], [125]) achieve improved accuracy by dynamically correcting wrong explanations.

While human-in-the-loop approaches that utilize high-level concepts represent an interesting research direction, researchers have pointed out their inherent limitations. For instance, [126] note that some image features other than the predicted concept may be responsible for a given prediction. Also, where there are high correlations between concepts, a particular concept may be implicated for a given prediction whereas in fact, the decision is caused by a correlated concept. Another important concern of this approach is that it requires a great deal of manual work to provide concept labels, and with large volumes of data it can be excessively laborious. Owing to this limitation, it is impractical to apply concept bottleneck modules on large-scale datasets containing millions of samples. However, effective techniques of mitigating this problem have yielded promising results. In ACE, Ghorbani et al. [127] propose to solve the problem by automating the annotation process. They achieve this goal by aggregating local patches in the input data into coherent and meaningful concepts with relevance to the network's prediction. Oikarinen et al. [122]) propose Label-free CBM that obviates the need for concept annotations for training. The authors accomplish this feat in four steps. First, GPT-3 [28] is employed to automatically generate the initial concepts which are then filtered to remove undesirable ones. Second, the concept activation matrix on the training dataset the training data is computed. The third step learns a projection needed to create a concept bottleneck layer. The last step involves learning the weights of the final network layer to make predictions.

Some Concept Bottleneck Models incorporate two tiers of human knowledge in their pipelines – human-defined concepts used in training; and direct human involvement in rectifying faulty concept predictions during inference. Another popular class of human-in-the-loop interpretability methods [128], [129], [130], [131]), known as eXplanatory Interractive Learning (XIL), employs human supervision to manually edit the heatmaps generated by conventional attribution methods like LIME, CAM Grad-CAM. Although Active Learning (AL) [132] also leverages human in the machine learning loop to improve performance, the fundamental difference is that XIL particularly focuses on achieving this goal by manipulating explanations. The intuition is that, by indicating areas in the saliency maps that are irrelevant to the given prediction, or important regions that have been missed by the attribution method, the network can learn to ignore spurious correlations in the data and produce more representative visualization maps. Usually, the expert-rectified heatmaps are presented to the model as additional annotated data in the spirit of data augmentation. Alternatively, the information from expert feedback may be incorporated into the XIL model through regularization with additional loss terms that seek to penalize deviations of the model's computed heatmaps from the human annotated ones. This line of works is also popular in other domains such as NLP [133] and multimodal multimodal vision-language frameworks [134].

**Knowledge graphs:** Knowledge graphs carry information in the form of entities, relations, and governing rules. Knowledge represented by this powerful data structure has proven effective when incorporated in machine learning frameworks to help generate explanations for model decisions. With knowledge graphs, explanations are easy to extract [135], [136] because of the rich semantic relations inherent in the representation. This kind of explainability is especially popular in applications such as product recommender systems (e.g., [19], [137], [138]) drug recommendations [139], [140]) and disease diagnosis [141], [142]).

In medical applications, especially in situations where data is limited, domain knowledge in the form of medical ontologies represented by knowledge graphs enhance both accuracy and explanainability of decisions. For instance, GRAM [143] incorporates an attention mechanism and ontological data to learn medical concepts in a framework that predicts and explains future onset of diseases from patients' medical history. In GRAM, ontological knowledge is used to learn the representations of medical codes, but this knowledge is not propagated directly to generate clinical visit embeddings and to contribute to final decision making. Yet, GRAM achieves significant improvement over the baseline in the insufficient data case, but does not take advantage of more data when available. Subsequent improvements (e.g., [144] leverage ontological knowledge in each step of the pipeline, and are, thus, able to exploit as much knowledge as provided by the data so as to boost performance.

**Logic rules:** As mentioned, some interpretability methods, notably conventional approaches that do not leverage domain knowledge, produce explanations that may be based on features irrelevant to the prediction, or may not be informative enough to capture the underlying rationale in a readily understandable form. Knowledge-driven solutions based on logical reasoning over symbolic rules help overcome both of these limitations. By enforcing compliance of explanations with prior knowledge about the tasks, rule-based methods [108], [145] ensure that only sound associations between predictions and explanations are allowed. Also, logic rules are unambiguous and are easy to understand by humans. Their strengths have therefore motivated state-of-the-art interpretability methods for deep learning.

From high-level concepts in the form of predicates, LENs [108] learns to provide easy-to-understand explanations using first-order logic statements. LENs has been proposed in two variants; as a post hoc module providing explanations for black-box frameworks, and as an inherent explainability model trained end-to-end. One major limitation of approaches based on logical rules is the requirement of the inputs to be symbolic predicates, a requirement which is difficult to meet in application domains that cannot utilize text-based data.

### 3.3 Knowledge-informed zero-shot learning approaches

**High-level concepts and logic rules:** When presented with given object categories and explanations on their visual appearances, humans can seamlessly extend this knowledge to unseen categories if additional descriptions of semantic relationships to the known classes are provided. For example, by describing a tiger as a big cat with black and yellow strips, it is easy for a reasoning agent familiar with the domestic cat to recognize a tiger when it is encountered for the first time. Many knowledge-informed zero-shot learning methods (e.g., [146] have exploited this idea to define visual concepts from text descriptions, and have achieved impressive results by reasoning over plausible concepts. The method is powerful as it can exploit the compositionality of objects with respect to known concepts, and logical rules can then be used to combine these concepts to match object descriptions of seen and unseen classes. For instance, [147] employ Logic Tensor Networks (LTNs) to achieve impressive performance in generalized zero-shot learning by adapting the relationship representations of concepts to distance measures, where closer relations are represented by shorter distances. Similarly, CLORE [146] extracts logical rules from textual explanations on images to capture high-level image attributes or concepts. This representation facilitates zero-shot image classification through logical reasoning over plausible concepts.

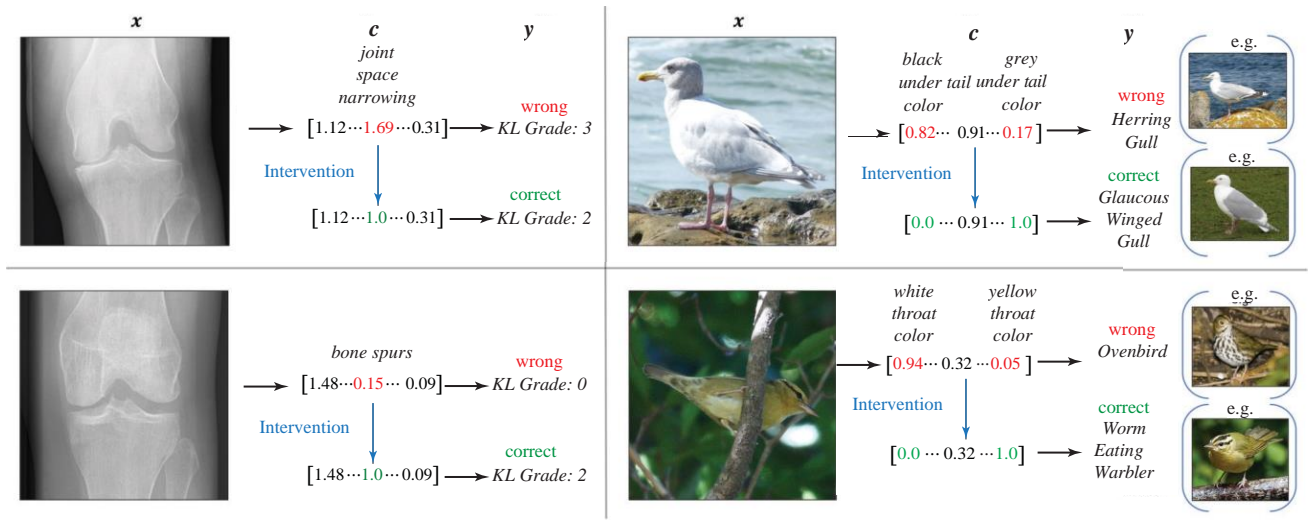

Figure 9. By manually editing concept predictions, overall accuracy can be improved [118].

Establishing the relation between entities without explicit examples in the training set is a challenging task. Yet, Li [148] show that logic rules can be used to establish the connection between unseen relations and some observed relations from a knowledge graph. The rules are extracted directly from the knowledge graph embedding, followed by rule-guided learning to extract connections observed and new relations. Rule learning over knowledge graph framework avoids the need for relation descriptions to understand the connection between seen and unseen cases.

**Compositional part-based reasoning:** Besides their effectiveness in adversarial defenses, compositional part-based reasoning methods have also been applied successfully in few-shot object recognition. In CORL (short for Recognizing Object by Components), He et al. [149] represent object categories with the knowledge of shared components and their spatial topologies. The knowledge base for training consists of two separate dictionaries, one for object parts and the other for their common spatial patterns. Trainable attention is also employed to reinforce object parts that are most relevant for recognizing each class. At test-time, the knowledge learned helps to recognize categories from just a few examples seen during training. Recognition as Part Composition (RPC), proposed by Mishra et al. [150] also achieves zero-shot and few-shot generalization by representing objects as constituent parts which are further decomposed into smaller units based on human human-level understanding. Besides zero-shot generalization, both CORL and RPC facilitate interpretability by virtue of the human understandable part relationships used for prediction. Additionally, RPC has shown adversarial robustness in object recognition tasks. It accomplishes this by being sensitive to improbable part compositions in the input. Successful implementation of part-based reasoning requires a careful choice of parts. For example, the parts used must not undergo sever or unrecognizable deformations. This could seriously affect performance. If necessary, it is possible to include some of the common transformations in the training phase, so that the model learns to make appropriate decisions and explanations during inference.

**Knowledge graphs:** Zero-shot knowledge graph completion methods are designed to recover unseen relations that emerge at inference time without any association to existing triples encountered during training. Some works [151], [152] try to predict new entities inductively by aggregating their neighbors. Although LAN [152] improves this general philosophy of enhancing neighborhood aggregation with attention mechanism that utilizes logic rules to estimate attention weights, the approach still does not generalize to new target knowledge graph embeddings. Therefore, to handle different entity types that might emerge with new knowledge graphs, some more recent studies (e.g., [153]) propose to transfer entity-independent knowledge from seen to target domains through meta-learning. Another approach relies on utilizing generative adversarial networks (GANs) to infer valid relation embeddings from text descriptions (e.g., [154], [155]) or ontologies [156], [157]. Generally, zero- and fewshot learning using knowledge graphs has made significant progress. It can be utilized effectively by models to make predictions over data that is not directly encoded in the graph. However, since most knowledge graphs usually capture knowledge in a specific domain, extending these frameworks to open-world generalization tasks is challenging. For this reason, large-scale pre-trained models which encode a much wider scope of world knowledge are employed to solve domain-agnostic generalization problems.

**Pre-trained foundation models and knowledge graphs:** Another important area where knowledge-augmented deep learning has achieved tremendous performance is zero-shot and few-shot generalization leveraging implicit knowledge encoded in pre-trained foundation models. Since generalization of machine learning systems is usually limited by the quantity and representative quality of training data, researchers propose to leverage the enormous multimedia data (usually, text and images) to improve generalization. The data is mostly obtained from online knowledgebases like Wikipedia, Google search results and news platforms. Pre-trained foundation models are particularly useful in natural language processing, computer vision and joint language-vision tasks. They can broadly be categorized into three main groups according to their application domain:

1) Large Language Models (LLMs) are used in natural language processing domains for language modelling or text generation. Prominent examples of LLMs include the Bidirectional Encoder Representations from Transformers (BERT) [26], PaLM [27], Generative Pre-trained Transformer (GPT) series like GPT [158], GPT-3 [28] and InstructGPT [29]. Typical applications LLMs include question answering (QA), a task which aims to train models on textual data to answer questions in natural language. Chatbots and conversational agents ChatGPT are examples of QA systems.

2) Pre-trained vision models: these are commonly used in open-world image classification, object detection and image segmentation. Popular models in this category include Florence [159] DALL-E [160] and Segment Anything Model (SAM) [161].

3) Multimodal Vision-Language frameworks – e.g., VL-BERT [162] and Contrastive Language-Image Pre-training (CLIP) [163] are focused on multimodal tasks such as visual question answering (VQA), where the goal is to train neural networks on images and text so that they can answer questions about visual scenes in natural language.

The general structure of a typical pre-trained foundation model, CLIP [163], is presented in Figure 10. It consists of three main processing stages. First, feature encoders are designed to extract textual and image features; second, a linear classifier creates a database of correct image-text pairs; finally, during inference, the model retrieves textual descriptions or caption that match given images.

Owing to the rich information contained in the massive training data, state-of-the-art foundation models implicitly acquire a vast amount of world knowledge and have shown incredible abilities to generalize to new tasks and data distributions. Studies have shown that, even out-of-the-box, pre-trained foundation models excel in few-shot and zero-shot generalization. For instance, Brown et al. [28] demonstrate that GPT-3 can perform diverse reasoning tasks when presented with a few in-context examples in the form of text prompts. Also, in the original study, Radford et al. [163] show through extensive comparison tests involving over 50 state-of-the-art models and 30 datasets that CLIP's zero-shot image retrieval performance is competitive even with supervised baselines. Similarly, with just a single prompt, [164] show impressive zero-shot reasoning capability of PaLM [27] and InstructGPT [29]. Multiple studies have therefore leveraged this implicit knowledge as a stand-alone entity (e.g., [165] or to augment explicit knowledge represented in various forms, like knowledge graphs [20] to achieve zero- or few-shot generalization in downstream tasks.

The tasks that have mostly benefited from the zero- and few-shot generalization ability of pre-trained foundation models are question answering (QA) and visual question answering (VQA). The QA task aims to equip machines with the ability to automatically answer natural language quesions by extracting or generating appropriate text in response to questions. This capability is useful in applications requiring automatic response to human queries, such as customer support and therapy chatbots. VQA aims to equip machines to be able to provide answers in response to natural language questions on an image. VQA systems can aid visually impaired persons to understand the content of scenes from camera dada.

Commonsense QA and VQA tasks require external knowledge about the question or image content. However, despite empirical evidence showing impressive performance of foundation models in these domains, out-of-the-box, these models typically lack domain-specific knowledge or commonsense reasoning ability since they cannot capture facts in context. Moreover, the predominantly web-based multi- media corpora used to train foundation models do not contain adequate examples of some kinds of data while unduly overemphasizing others. Specifically, this information is subject to reporting bias, a situation where some rare events and facts are disproportionately represented at the expense of more common and relevant everyday facts. For example, compared to mundane activities like cleaning the kitchen sink, it is more likely for a school shooting incident to assume prominence online. Furthermore, some kinds of data are mostly accessible only in specialized domains. Yet, a deep learning model trained in an unsupervised manner to glean general knowledge requires sufficient number of examples on each of these cases to adequately capture relevant information required to achieve acceptable performance. Unsurprisingly, the developers of CLIP in their report [163], have bemoaned the model's uncharacteristically poor performance on "complex, or abstract tasks" like satellite image classification and tumor detection. However, it must be noted that the domain-specificity—rather than any inherent "complexity" or "abstractness" of these tasks—is to blame for this subpar performance. To address this limitation in downstream tasks, various techniques exploit commonsense or domain-specific knowledge encoded in knowledge graphs to augment pre-trained foundation models. The general workflow of this line of approaches is explained by Figure 11.

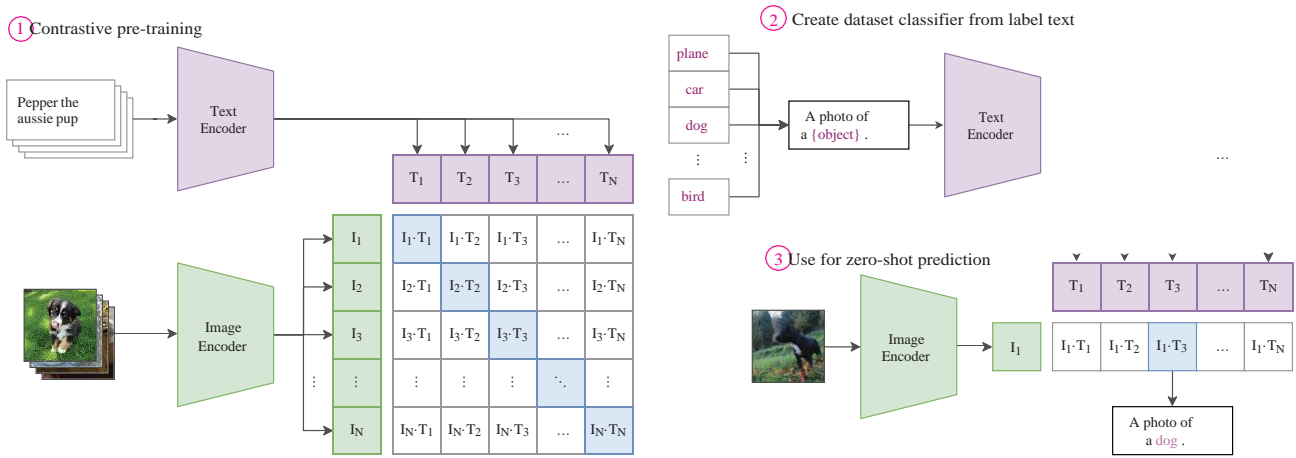

Figure 10. General procedure of training CLIP [163] for inference.

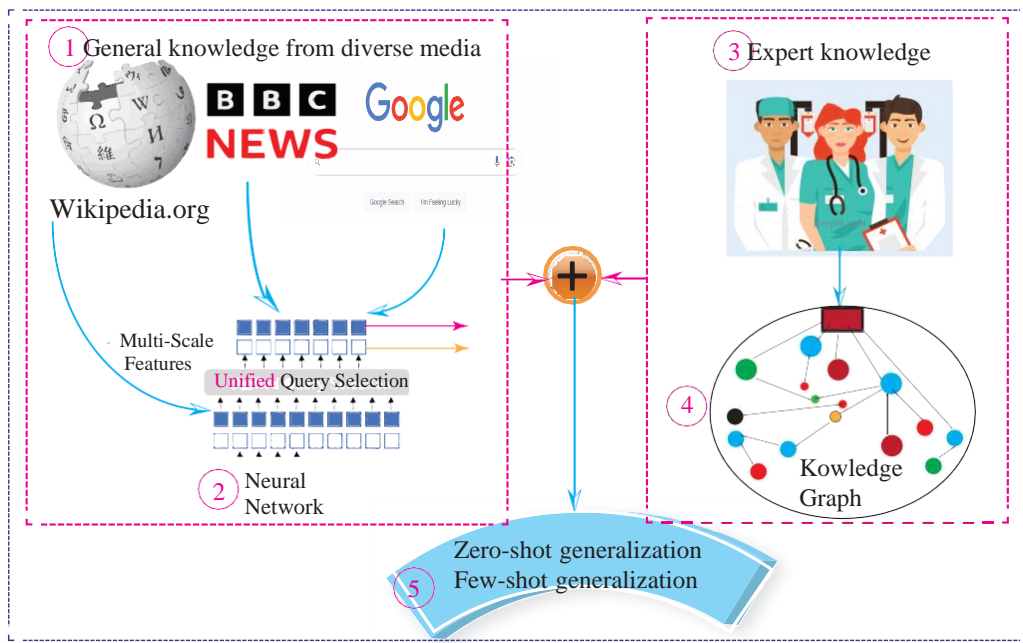

Figure 11. General workflow of integrating knowledge graph and pre-trained model for zero- and few-shot learning: Information from various online sources (1) is used to pre-train a neural network (2) which acquires broad world knowledge. On the other hand, domain-specific expert knowledge (3) is encoded in a knowledge (4). The two knowledge systems (2) and (4) then complement each other in the end task (5).

An important body of work seeks to augment pre-trained foundation models with explicit knowledge from content-rich knowledge graphs like ConceptNet [110], WordNet [166] and ATOMIC [167]. To capture contextual commonsense knowledge from these sources, these methods [168], [169] propose to fine-tune pretrained BERT [26] to commonsense knowledge graphs using the text phrases that represent graph nodes. Bosselut et al. [168] first develop a generative transformer model named COMmonsEnse Transformer (COMET), and then use it for knowledge graph completion by re-training the pre-trained LLM on the existing set of knowledge tuples. The LLM learns the knowledge graph structure using the example tuples and then generates compatible tuples to complete the graph Malaviya et al. [169] transfer implicit knowledge from the pre-trained large language model using a masked language modeling loss to facilitate contextual knowledge graph completion.

The approach allows the vast amounts of implicit knowledge contained in powerful foundation models to be transferred to explicit representations in knowledge graphs. The method achieves improved results over BERT-only representations. However, in practical QA tasks, a pre-populated knowledge graph may not adequately capture the relevant context about a given question. To solve this commonsense zero-shot question answering problem, Bosselut et al. [168] employ the COMmonsEnse Transformer (COMET) [168] — which is based on the generative pre-trained language model GPT [158] — to dynamically generate context-appropriate knowledge graph triples in response to each question posed. Presented with a question and a corresponding context as a root node, COMET generates intermediate inferences as child nodes that connect the given context to the answer choices which act as leaf nodes of the graph. Finally, probabilistic inference over the graph retrieves the most suitable answer. A common weakness of knowledge graph augmented learning is the limited coverage or scope of world knowledge encoded by these knowledge bases. This limitation might also lead to irrelevant knowledge being retrieved from a knowledge graph when the precise information is not captured by the graph. To improve coverage of commonsense facts across a variety of domains, KRISP [170], combines information from multiple knowledge graphs to augment pre-trained BERT for open-domain VQA. This design expands the range of application of the model beyond a single domain.

Other works try to overcome the problem of restricted scope or coverage of encoded knowledge –and hence, incidences of irrelevant knowledge retrieval – of publicly available knowledge graphs by leveraging pre-trained (LLMs such as) GPT-3 instead. Rather than utilizing explicit knowledge from structured knowledge graphs, some recent few-shot VQA methods such as PICa [165], KAT [171], Prophet [172] propose to leverage implicit and unstructured world knowledge from GPT-3 [28] for enhancing visual question answering. At test time, when required to answer a querie on an image-question pairs, PICa converts the images into corresponding captions. These captions and accompanying questions, along with a small sample of n-context VQA examples are then fed into the pre-trained language model as textual prompts to help retrieve relevant knowledge for making predictions. However, as GPT-3-only predictions may lack domain-specificity, KAT proposes to combine explicit knowledge from Wikidata [173] obtained through contrastive-learning, and implicit knowledge from GPT-3 to facilitate versatile/multipurpose but domain-appropriate predictions.

On the other hand, by noting that captions automatically generated from images may be limited in capturing all relevant scene information and semantic relationships required for reasoning-based VQA, Prophet [172] first learns relevant answer heuristics which are then included in the prompts to enhance the prediction task. With richer and more task-specific information for answer prediction, Prophet surpasses existing state-of-the-art models in accuracy.

Foundation models are extremely large deep neural networks that take large amount of memory space and processing times. They typically require high-end systems to operate. Therefore, in practical applications, the demand on computational resources is a major limitation of models that directly utilize large-scale pre-trained models. Therefore, to address this problem for resource constraint zero-shot and few-shot visual question answering applications, more compact and efficient frameworks — e.g., VLC-BERT [174] — which uses the COMmonsensE Transformer (COMET) to generate and integrate contextualized Knowledge into a pre-trained vision-Language model based on VL-BERT [162] — has achieved impressive results. With ap-proximately 118 M parameters, VLC-BERT achieves competitive accuracy with large-scale networks augmented with knowledge from pre-trained language models like GPT-3, whose parameter count is over 175 billion.

Zero-shot and few-shot generalization in QA and VQA based on foundation models have reached a matured level in terms of performance. The main challenges that remain to be addressed is the interpretability of these models. Owing to their large sizes and the wide scope of data they are trained on, it is difficult to interpret them at the decision level. Other important issues concern their use or misuse. Owing to their extraordinary power and immense versatility, they can be used for malicious purposes not intended by design, thus, raising privacy and ethical concerns. As these models are exposed to unrestricted store of information, the expressions they produce in response to questions need to respect social boundaries. Surprisingly, as at this point, much of the attention is still on improving test-time performance. It is however anticipated that when performances begin to plateau as most of the outstanding accuracy-restraining problems are addressed, attention would quickly switch to these other important issues.

## 3.4 Summary of the main features of prior knowledge-informed approaches

Table 1 summarizes the major performance characteristics of the common knowledge-augmented deep learning frameworks. As shown, some of the knowledge-informed methods simultaneously address multiple objectives such as accuracy, explainability, adversarial robustness and zero- or few-shot generalization. In general, methods that leverage compositional part relationships excel at almost all four objectives. On the other hand, some approaches may enhance their main objective while compromising other figures of merit. The interpretability-accuracy trade-off – which still a subject of intense debate anyway (see [130], [131], [175], [176], [177]) — associated with some knowledge-informed techniques is a classic example of this situation.

Explicit mathematical equations and physics-informed methods can enhance neural networks by leveraging the underlying knowledge to constrain the input or output space, or even the architecture of the model. These constraints can often lead to overall accuracy reduction, although the fidelity and conformance of the decisions with physical environments would improve. Other approaches aim to exploit mathematical representations to generate entirely new training datasets when none is available, or augment existing ones using synthetic data augmentation approaches. This class of methods invariably lead to improved accuracy and robustness. It should be noted that the accuracy improvement results presented are not meant to serve as a basis for comparing the various methods. Given that the tests are carried out under different settings and on different datasets, and since the tasks themselves exhibit varying degrees of difficulty, the performance figures are not relevant for comparison purposes. The information only explains which models achieve accuracy improvements over specified baselines. Unless otherwise stated, these baselines are the equivalent model configurations that do not leverage prior or domain knowledge.

Table 1

Performance characteristics of the common knowledge-informed methods. The arrow symbols ↑ and ↓ indicate improvement (↑) and a drop (↓) in standard or clean accuracy (Acc.), respectively, with respect to the baseline while ↔ indicates no appreciable change. X and √ indicate whether the given model possesses interpretability or explainability (XAI), adversarial robustness (Adv) and zero- or few-shot learning (ZSL/FSL) capabilities. The presented accuracy values are reported for the settings that yielded the best results relative to the baseline.

| Model | Knowledge representation | Objective | | | | Highlights |
|---|---|---|---|---|---|---|
| | | Acc. | XAI | Adv. | ZSL/FSL | |
| ROCK [117] | Compositional parts | ↔ | ✗ | √ | ✗ | * ROCK's clean accuracies are competitive with the most effective methods that recognize object as-a-whole. |
| CORL [149] | Compositional parts | NA** | √ | ✗ | √ | Part-based reasoning methods are capable of achieving multiple objectives (adversarial robustness, interpretability and ZSL or FSL) simultaneously. NA **Empirical results on clean accuracies are not available. |
| RPC [150] | Compositional parts/ concepts | NA** | √ | √ | √ | RPC decompose images into two hierarchies – salient parts constituent concepts. NA **Empirical results on clean accuracies are not available. |
| MIDPhyNet [14] | Physics-informed modelling | ↑ 15% | ✗ | ✗ | ✗ | Here, a neural network is trained using intrinsic mode functions which are formed by decomposing a physics-based model of the system under investigation. |
| CGIntrinsics [178] | Physics-based rendering | ↑12.1% | ✗ | ✗ | ✗ | Uses physical laws to render photorealistic scenes that provides additional training images as a form of data augmentation. In addition to test accuracy, these methods usually improve general (but not adversarial) robustness. |
| ACnet [179] | Math equations | ↓ | ✗ | ✗ | ✗ | ACnet (Architecture-Constrained network) introduces math-guided constraints that, once satisfied, causes the model to obey underlying physical laws. This however, leads to a marginal drop in accuracy. |
| GLGExplainer [180] | KG/rules | NA | √ | ✗ | ✗ | Employs logic rules to provide global explanations as logical combinations of learned graphical concepts. |
| CBM [118] | CBM/ HITL | ↑ 14.5%** | √ | ✗ | ✗ | By allowing expert intervention on concept predictions, CBMs consistently report simultaneous gains in explainability and final prediction accuracies by clear margins. This is despite intense debate ( [131], [175]) about the explainability-accuracy trade-off. |
| CBM-AUC [181] | CBM / HITL | 4.3%** | √ | ✗ | ✗ | ** The reported value is the accuracy (mean F1 scores over all predictions) improvement relative to the original CBM [118]. |
| DL-AMC [15] | CBM / HITL | ↑3.51% | √ | ✗ | √ | Accuracy improvement is relative to a baseline that does not employ CBM / HITL. |
| CEMs [182] | CBM /HITL | ≈5% | √ | ✗ | ✗ | Accuracy improvement is relative to a baseline that does not employ CBM / HITL. |
| CB2M [183] | CBM / HITL | 6.3%** | √ | ✗ | ✗ | To improve HITL-based interventions, CB2M employs memory which keeps track of the outcome of past interventions, allowing mistakes to be detected refinements to be made. * Improvement relative to CBM [118] |
| CAIPI [128] | XIL / HITL | ↑ 37.6 % | √ | ✗ | ✗ | CAIPI uses human feedback on incorrect explanations to create additional data (counterexamples) for training and shows that prudent data augmentation can achieve dual-interpretability and accuracy goals. |
| eXBL [130] | XIL / HITL | ↓ | √ | ✗ | ✗ | Consistent with the explainability-accuracy trade-off, most XIL frameworks, including eXBL, sacrifice accuracy for interpretability. |
| Ref. [131] | XIL / HITL | ↓ | √ | ✗ | ✗ | Employs techniques that use filters to refine output based on ground-truth highlighted regions and Grad-CAM predictions. Tested another technique that uses regions of interest (ROIs) maps to direct the networks attention to expert-provided annotations. All methods show improved interpretability. |
| Ref. [184] | XIL / HITL | ↔ | √ | ✗ | ✗ | CAIPI in Practice [extension of CAIPI [128] that focuses on data efficiency, allows human experts to edit explanations and also correct wrong predictions–resulting in competitive accuracy to be reached with fewer training examples. |
| KGIN [185] | KG | ↑14.5% | ✗ | ✗ | ✗ | KGIN models user intents as attentive combinations of KG relations. |
| Ref. [107] | Logic rules | ↑2.3%/1.1%* | ✗ | √ | ✗ | Logic rules can simultaneously boost accuracy along with a combination of other objectives (e.g., explainability and adversarial robustness), especially when combined with other knowledge types like KG and probabilistic representations. |
| LOGICDEF [109] | Logic rules/KG | ↑55% | √ | √ | ✗ | LOGICDEF uses rules to provide explanations as to why a system is attacked and this inspires measures to overcome subsequent adversarial cases. |
| ExCAR [186] | Logic rules/ probabilistic | ↑ 6% | √ | ✗ | ✗ | ExCAR integrates representation learning with structure learning of logical rules. |
| KEMLP [187] | Probabilistic/ Logic rules | ↔ | ✗ | √ | ✗ | KEMLP achieves high adversarial robustness via integrating domain knowledge represented by logical relationships into probabilistic models. |
| S-AL [188] | Ontologies | NA | √ | √ | ✗ | S-AL leverages ontological information to improve the handling of relations among classes. |
| OntoPrompt [189] | KG/ Ontologies | NA* | ✗ | ✗ | √ | Knowledge graph-based methods like OntoPrompt are effective in zero- and few-shot learning settings when used alone or in combination with other constructs. NA* – No accuracy baselines are available. |
| CA-ZSL [190] | KG | NA* | ✗ | ✗ | √ | Exploits geometric relationships among known and unknown objects in an image for zero-shot inference. NA* – No accuracy baselines are available. |
| DGP [191] | KG | NA* | ✗ | ✗ | √ | Proposes a Dense Graph Propagation (DGP) module to connect distant nodes of a knowledge graph in a graph convolutional neural network framework. NA* – No accuracy baselines are available. |
| GRAN [192] | KG | NA* | ✗ | ✗ | √ | GRAN employs graph modeling to leverage visual characteristics and semantic relations of multiple objects in a scene. NA* – No accuracy baselines are available. |
| Prophet [172] | GPT-3 /KG | NA* | ✗ | ✗ | √ | Combining implicit knowledge from pre-trained foundation models and explicit knowledge from KGs dramatically improves ZSL/FSL. NA* – No accuracy baselines are available. |
| KAT [171] | GPT-3 /KG | NA* | ✗ | ✗ | √ | Retrieves implicit knowledge from GPT-3 while using CLIP to extract explicit knowledge from knowledge bases. NA* – No accuracy baselines are available. |

# 4 IMPROVING DEEP LEARNING WITH COGNITIVE INSIGHTS

Knowledge-informed methods attempt to enhance deep learning by equipping models with domain knowledge available to humans. This enables deep learning systems to handle robustness and data-insufficiency problems better. The knowledge mainly entails how the world works. Motivated by the incredible power of the human mind, and how humans excel remarkably in these tasks, brain-inspired techniques that leverage principles of biological cognition are proposed to exploit how the mind works to solve these problems.

The design of cognitive architectures has follows three dominant approaches (see [193]) The first group of methods comprises symbolic cognitive architectures (e.g., EPIC [?], [194]; MusiCog [195] which rely on symbolic representations to encode real-world knowledge. These architectures then utilize predefined instructions or explicit rules to process the symbols in order to exploit encoded knowledge. This design makes them suitable for planning and logical reasoning tasks. Moreover, their use of explicit rules provides a clear view of their operating mechanisms and rationale for their decisions and actions. However, since the rules are explicitly defined and fixed, their scope is often very limited. Moreover, it is not easy to adapt them to unanticipated situations that arise in deployment. Therefore, symbolic cognitive architectures are the least suitable techniques for tasks that require adaptation through online learning. However, owing to their conciseness and grounding in first principles, symbolic representations are inherently transparent, making the reasoning process and consequent decisions of the cognitive architectures that adopt them highly interpretable.

The second group of approaches, known as connectionist or sub-symbolic architectures – e.g., ART [196]); MDB [197] –work similar to artificial neural networks. They are designed to process information like the brain by mimicking its conceptual organization with massively interconnected network of neurons (see, for example [198], [199]). Connectionist cognitive architectures may implement biological or artificial neuronal models, or an equivalent structural logic that facilitates learning. They acquire their knowledge by updating internal parameters or weights through a gradual process of interactions with their environments, thus mimicking a learning paradigm akin to biological cognitive process. A major problem with this category of cognitive systems is that the acquired knowledge directly maps perceptual inputs to final decisions in an opaque way, with no intermediate representation that can be readily interpreted by the model developer or the end user.

The final category, hybrid cognitive architectures, includes popular models like CARACaS [200], [201]; iCub [202], MIRIAM [203] and IMPACT [204]. They incorporate both symbolic and connectionist capabilities in their design to facilitate learning and adaptation, as well as reasoning over acquired knowledge. By internal organization, the two most common categories of hybrid architectures that can be identified are modular and integrated systems. While integrated methods attempt to encode different aspects of domain knowledge and solve other learning-related problems using a single complex network, approaches based on modular cognitive architectures utilize specialized submodules to accomplish this goal in a task-dependent manner. For example, CARACaS [200] employs a hybrid technique that incorporates a connectionist reinforcement learning unit to process s ensory data from multiple sources to acquire knowledge about the environment in a robust way to aid localization and navigation. All other functions like planning and behavior coordination are handled by a separate symbolic processing module. Whereas some hybrid systems like CARACaS [200] use separate submodules for their symbolic and sub-symbolic components, others like iCub [202] and IMPACT [204] propose an integrated approach that allows new skills to be incorporated into their symbolic modules. IMPACT [204], for instance, is an autonomous robot system that facilitates continual and open-ended learning by integrating the sub-symbolic reinforcement learning unit and the symbolic planning unit in such a way that the knowledge acquired through learning is automatically encoded in symbolic form. This feature allows the symbolic knowledge base, which drives the robot's actions and behaviors, to be extended continuously as it interacts with its environment.

## 4.1 Overcoming Adversarial Attacks with cognitive insights

### 4.1.1 Cognitive architectures for adversarial robustness

Although cognitive architectures are not usually designed with the specific aim of addressing adversarial attacks, their quest to ensure general intelligence through learning, perception and reasoning invariably leads to the implicit goal of improving adversarial robustness. It should be noted that adversarial attacks are usually crafted to exploit vulnerabilities of conventional deep learning methods, and since most of these attacks generally pose no problems to humans, it can be argued that enhancing deep learning with human-like reasoning and perception capabilities implicitly constitutes adversarial defense.

Only learning-capable cognitive architectures with the potential to acquire new knowledge online and adapt their behaviors have the ability to handle new threats that arise in the course of operation. Additionally, such an agent requires knowledge and reasoning ability to make the right decisions. Hybrid robotic systems like IMPACT [204] and iCub [202] are designed to meet these requirements.

The approach of the hybrid systems is consistent with the dual process theory [205], [206]) of biological cognition which stipulates that the brain deploys two separate systems for fast and slow processing of information. The fast and slow systems are also called system 1 and system 2, respectively. According to this theory, system 1 is responsible for fast or spontaneous decision making under the control of a subconscious thing while system 2 adopts a rational and systematic reasoning approach to solving problems. The roles played by the two systems complement each other to achieve the desired balance in accuracy of decisions and speed of response. In the biologically-inspired cognitive architecture implementations, the fast system relies on making predictions from knowledge acquired through learning by a connectionist system (e.g., a deep neural network). The slow component employs reasoning and planning ability of a symbolic processing module to solve difficult problems whose solutions are not explicitly encoded by the learning system. The operational logic of such a brain-inspired cognitive architecture is akin to real biological cognition which is known to deploy conscious and subconscious "minds" to solve problems requiring either analytical and systematic reasoning or spontaneous action.

As a matter of fact, humans' remarkable ability to recognize aberrant situations such as adversarial attacks is developed over a long period of accumulated experience gathered through life-long learning, and problem solving. This capacity is consolidated through repeated interactions with their environments. Using such a strategy, the SOFAI [207], [208] cognitive architecture employs a fast learner that continually improves by updating itself through the problem-solving experience of a slow reasoner. SOFAI uses a subsidiary meta-cognitive module to supervise the cognitive activities of the two main cognitive systems. The meta-cognitive module keeps track of the agent's state, percepts from the environment, and the ability of the fast system to produce desired solution under the given conditions. By default, system 1 handles any problem if it can provide a satisfactory solution, otherwise the meta-cognitive arbitrator assigns the task to system 2.

Another critical factor that accounts for the robustness of human perception is the ability to combine different sensory modalities (e.g., visual, auditory, tactile, olfactory) in a statistically optimal way to create a unified world model, see [209], for example, for further discussion on this concept. To arrive at a sound judgement about the environment, the brain processes different kinds of sensory information and combines them to form a coherent representation of the environment.

Two main principles of multimodal processing are commonly employed: multisensory cue integration and cue segregation. Integration of multisensory cues involves their joint processing to characterize an environmental property as a single unit. Bayesian cue integration or deep learning can be employed to combine the multisensory signals. Cue segregation processes each cue separately to produce independent representations of the target environmental property. Figure 12 shows these concepts with examples. Figure 12a) shows the conceptual framework of multisensory integration. A navigation robot with LiDAR sensor and a stereo camera both measure the distance to the two cars blocking the road. The estimated distances from the two sensors are integrated to produce a single distance value. In this way, the effect of corrupted input from one modality is minimized. Figure 12b) depicts the principle of sensory segregation. The navigation robot featuring camera and audio sensors wishes to understand the scene in front, i.e., whether the cars have crashed or stopped safely in close proximity. Assuming the cars suddenly appear in front of the robot but the visual signal is unclear to inform this judgement. In this scenario, the independently processed visual signal and auditory signal must both indicate a crash for such a conclusion to be made. In other words, if there is a crash, the two cars must come close enough (visual), and the incident must be accompanied by a loud sound (auditory).

Cue integration of multisensory perception is popular in cognitive architectures, with most utilizing neural networks for integration. For example, in the DAC-h3 cognitive robot, Moulin-Frier et al. [210] employs an adaptive learner to learn the representations of multisensory inputs that include tactile, visual and linguistic modalities. This process is followed by data association and alignment to transform the resulting information into a unified percept. Also, the MDB cognitive architecture [197] employs a combination of sensors whose signal association is learned by a neural network while a genetic algorithm performs data alignment. Sometimes information from multiple sensors can be processed and handled separately, with neither integration nor segregation performed. It is possible, in this case, to use one modality for primary perception while another serves as back-up in case of failure of the main percept.

In general, using multimodal information allows the reasoning system to improve robustness and overcome challenges that may arise when one sensory modality produces degraded or flawed outputs as a result of component failure, adversarial attack, or noise in the sensory data. However, some sensors may negatively impact each other when used in combination, thus, negating the intended benefit of multisensory perception. For instance, electromagnetic interference from inertial sensors and analog indicators may hamper performance if proper care is not taken in their joint use.

### 4.1.2 Brain-inspired DNNs for adversarial robustness

Despite the impressive learning ability of deep neural networks, they are uncharacteristically sensitive to subtle, context-irrelevant changes in their input data. Consequently, state-of-the-art neural networks achieve human-like prediction accuracy in tasks like image classification under suitable conditions but there is still a large performance gap between these deep learning models and humans in terms of adversarial and general robustness. Furthermore, notwithstanding their often-touted biological motivation and organizational similarity, the analogy between current convolutional neural networks (CNNs) and the human brain is only valid at a crude conceptual level.

Unsurprisingly, some of the attempts seeking to close the robustness gap have focused on improving the biological realism of deep convolutional neural network models. By incorporating a CNN layer with neuroanatomical constraints that simulate the primate primary visual cortex (or V1), [36] achieve a remarkable improvement in adversarial robustness on image classification. Their network, named VOneNet, incorporates a model of the primate V1 in the initial block of layers, followed by a block of conventional CNN layers. By implementing convolution operations on the input data with Gabor filter bank of high and low spatial frequency filters (SFFs), the first layer within the V1 block have their weights constrained to produce similar output to the response of the equivalent component of the primary visual cortex. The following layer approximates cell nonlinearities using spectral and rectified linear transformations while the final layer simulates neuronal stochasticity by affine transformation and the injection of Gaussian noise. Interestingly, the model demonstrates high level of immunity against a wide range of adversarial and common perturbations, outperforming several conventional adversarial defenses. It also achieves a competitive level of clean accuracy, thus, demonstrating a clear advantage over most conventional adversarial defenses which usually incur a severe penalty on clean images.

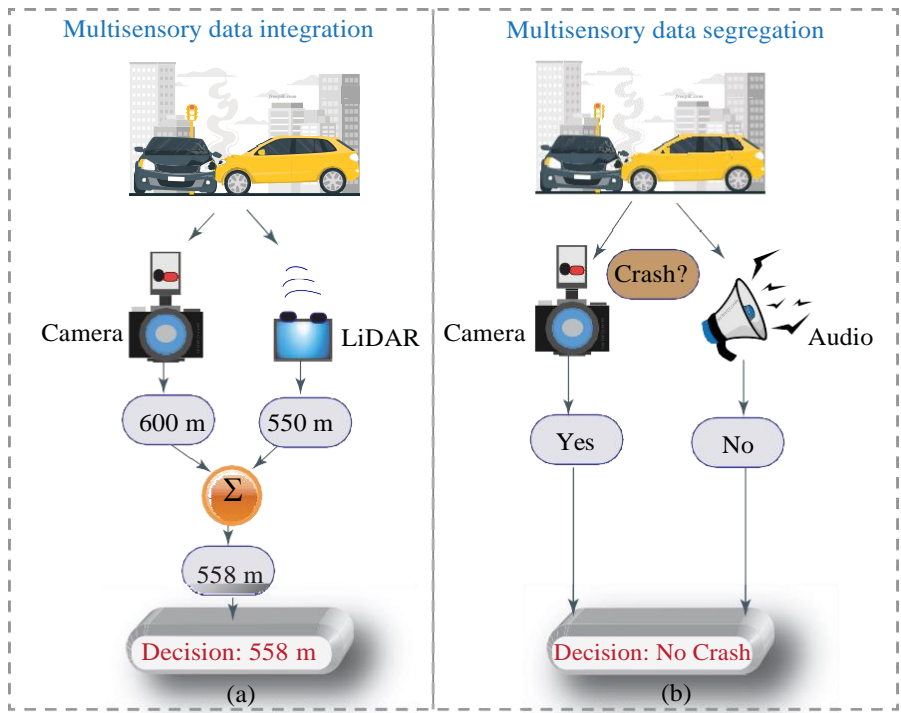

Figure 12. Two widely utilized principles of multimodal perception: multisensory data integration and multisensory data segregation.

Furthermore, by manipulating various components of the V1 block, the authors make very interesting observations with potential ramifi- cations for the understanding of adversarial robustness in neuroscience domains. First, when any of the components is removed or altered, the resulting model reduces in adversarial robustness, but still outperforms the baseline without the biological V1 component. This implies that all components in the primate V1 may work synergistically to achieve robustness. They also discover that some of the components of the V1 may respond differently to different types of adversarial and common image corruptions. For instance, removing high frequency Gabor filters in the first layer lowers robustness to white box attacks while increasing robustness to artifacts like image blur. Finally, incorporating neuronal stochasticity only at training time has more pronounced impact on adversarial robustness than employing this feature only during inference.

Baidya et al. [211] extend the VOneNet framework to incorporate an ensemble of multiple primary visual cortex models, each of which is tailored to mitigate a specific adversarial or corruption type. This modification results in improved robustness against a wider range of adversarial and common corruptions, suggesting that different V1 models may possess attributes relevant to solving different adversarial problems and can complement one another when used simultaneously. Dapello et al. [36] further extend their original work to demonstrate that functional simulation of late stage visual representations has a similarly beneficial effect on robustness. Specifically, the show that aligning the representations of a CNN and the Macaque Inferior Temporal (IT) cortex by enforcing similarity constraint between the biological and corresponding CNN components instills human-like recognition attributes into the model, including high adversarial robustness and human behavioral error patterns for unseen object classes.

In addition to incorporating high-level neuroanatomical concepts into deep learning model architectures, biological alignment has also been pursued from the perspective of task-aligned training. For example, using a conventional CNN, [212] propose to transfer the robustness of the visual system of the macaque monkey to the artificial neural network by jointly training their so-called MTL-Monkey DNN model on ImageNet dataset [213] and neural response data from the recordings of the primary visual cortex of macaque. During training, the network was required to concurrently classify an ImageNet instance and categorize the neural activity of a macaque monkey viewing the same image. The goal of this multi-task learning was to optimize the network on the individual task objectives, leading to good performance on both tasks, and yielding common representations that can be exploited by each task. To better understand the rationale behind the robustness of this class of brain-inspired networks, the authors leverage saliency visualization maps that explain the robustness characteristics of the model. Specifically, they observe that the jointly trained brain-inspired networks tend to focus more on salient image regions than noise or irrelevant perturbations, just like the primate visual cortex. These results show that training-level alignment can help transfer neuro-acuity from the brain to deep learning models, and equip them to improve on robustness and other performance characteristics that humans currently handle better than artificial intelligence systems.

More direct evidence of the utility of task-aligned neural networks in explaining human perception may be provided by recent studies [214], [215]) which demonstrate important similarities in the behavioral traits of DNNs and human or animal subjects in visual perception tasks. In particular, [214] show that DNNs trained on face recognition tasks suffer the face inversion effect [216]) – the dramatic drop in recognition accuracy for inverted faces (i.e., faces turned upside down) compared to upright faces – that also characterizes human face recognition. However, the same CNNs did not show this effect when trained on different but related tasks like generic image classification, even when the image categories include human faces. That is, the only setting that produces sensitivity to face orientation is the one in which the network is trained to recognize the identity of the faces involved (but not when the task is merely trying to tell whether the object in question is a human face or something else). Therefore, the precise task requirement is to blame rather than any representational peculiarity of the faces themselves. To further show that the face inversion phenomenon is not specific to faces, they trained their CNN on car model classification dataset and it achieves inferior accuracy on inverted cars compared to upright ones. Finally, training the network on inverted cars resulted in better performance on same. Therefore, the factors responsible for this behavioral anomaly can be attributed to task-specific optimizations developed through evolution and the fact that faces encountered in natural scenes are usually upright. These findings explain the real reasons for the observed phenomenon of face inversion effect, thus, challenging the longstanding assumptions that face recognition exhibits "special" intrinsic properties [217], [218]) not shared by psychological processes concerned with the recognition of other objects.

Methods that achieve representational alignment between neural networks and humans through similarity judgement experiments [219], [220] have also shown that human-alignment improves adversarial robustness and few-short generalization. Alignment in this manner allows neural networks to acquire effective representational capabilities, including the use of rich semantic information and global constraints. However, Sucholutsky and Griffiths [221] notice a U-shaped dependence between the performance and the degree of a neural network's alignment with human representations. Specifically, moderately-aligned models are less adversarially robust, and are also poorer few-shot learners compared to their weakly- and highly-aligned counterparts. Moreover, the alignment dynamics are highly influenced by the task or dataset used in the study. This suggests that neural networks must be optimized to a high degree of human-alignment to guarantee the benefits of human-like performance characteristics. When this cannot be ensured, empirical test should be conducted to establish that the level of alignment achieved is not within the "moderate" range and, hence, is not producing diminishing returns on adversarial and few-shot generalization performance.

## 4.2 Improving interpretability with cognitive insights

It is interesting to note that the lack of analytical support for the decisions of deep learning systems is not currently among the major problems brain-inspired artificial cognitive systems seek to address. In fact, it can be argued that the need for explainability is driven by the awareness that artificial intelligence currently lacks the level of intelligence needed to warrant unconditional trust by humans. If accuracy and robustness ultimately improve across many tasks and domains, trust will as well increase. After all, humans do not necessarily need to explain their decisions and actions. Therefore, it is clear why more effort on cognitive architectures and brain-inspired neural networks is rather vested in bestowing human-like intelligence capabilities to AI systems. This notwithstanding, in applications like human-robot interaction (HRI), mutual understanding and effective collaboration can be enhanced by ensuring the agent can make decisions and communicate in ways consistent with human subjects.

### 4.2.1 Improving interpretability with cognitive architectures

In cognitive architectures, learning modules represent and use their knowledge in a non-intuitive way. Therefore, decisions made by these modules cannot be understood by humans. On the other hand, symbolic architectures incorporate well-defined rules or instructions that guide behavior, making the resulting intelligent agents' actions readily interpretable. Cognitive architectures that utilize the dual learning and symbolic reasoning frameworks can easily provide explanations for the decisions of their reasoning modules if required. For instance, Augello et al. [222] propose a conceptual explainability framework for a hybrid cognitive architecture used in a robotic system for gesture recognition. The approach relies on the decision logic provided by system 2 (i.e., the reasoning module) to derive plausible explanations that characterize the categorization of gestures by the social robot. Unfortunately, since this hybrid architecture employs systems 1 and 2 alternatively, it follows that not all decisions can be explained by such a design. Specifically, the inherent opacity of the learning module precludes explanation, and whenever this module is engaged the decisions taken cannot be interpreted.

An alternative approach to explainability is utilized by Wu et al. [223], who proposed a hybrid cognitive architecture for autonomous vehicles with the aim of achieving interpretability and improved learning capacity at the same time. The model incorporates Markov logic network (MLN) for symbolic representation of domain knowledge and a reinforcement learning module, deep Q-network (DQN), as the connectionist module. The DQN is responsible for learning driving policies while the MLN learns logic rules for evaluating driving actions. Thus, the MLN provides interpretable reasoning-based decision-making that abstracts the opaque low-level decision process of the connectionist module. Here too, interpretability is only partial, since the lower decision-making layer remains opaque. Another useful approach to tackling the interpretability problem, used in CARACaS [200], solves the problem by utilizing a decision tree to generate logic rules from the model's decisions and actions. These rules are specifically meant for explaining the underlying rationale behind the given decisions. In their design, the decision tree has access to all the relevant sensory information that triggered the given decision. Therefore, the generated rules are capable of capturing a sufficient amount of detail since that explains the agent's behavior.

A special form of interpretability, called cognitive salience in reference to the gradient-based saliency visualization methods used in image classification (described in Section 2.2 of this paper), is devised by Cranford et al. [224] based on cognitive modelling to explain how humans make decisions in a cybersecurity setting. Specifically, using the ACT-R cognitive architecture, the authors model how individuals weigh different features based on past experience on reward and penalty to make decisions on cybersecurity attacks. In this implementation, interpretation of feature importance from the ACT-R model is provided by cognitive salience, which is computed by taking the derivative of an interpolation equation that measures the retrieval of outcomes from memory. One of the most important behavioral traits discovered by the model is that instead of using all available information, humans usually make decisions by relying on past experiences, even when these decisions are irrational. The results show that interpretability derived from cognitive architectures may be useful in explaining psychological processes in the brain.

### 4.2.2 Improving interpretability with biologically-informed neural networks

Biologically-informed neural networks are inherently explainable neural network models that are designed to approximate the underlying structure and dynamics of real biological systems that they imitate. Like biological cognitive entities, these models are generally more capable of capturing the actual relationships of biological entities (input signals) and their interactions compared to conventional DNN models. Here, network components may have precise biological associations and connotations. For instance, nodes and their interconnections may represent gene ontology terms or biological pathways. One of the pioneering works on biologically-informed neural network is DCell [225], which is designed to facilitate effective cellular growth prediction by exploiting the hierarchical structure of gene ontologies. The structure of the network allows the progression of information flow from input genotype to phenotype response to be presented in a way that mimics the actual biological mechanism, making it an intuitively interpretable proxy for the real biological system.

In cases were the underlying structures or the dependencies of biological entities are deemed too complex to be fully represented in a truly transparent form, additional workarounds are employed. For instance, in a study to predict transcription factors (TFs) responsible for inducing specific transcriptional changes (targets) in gene expression during a disease, Magnusson et al. [226] design an interpretable network that does not seek to completely mimic biological gene regulatory networks from architectural point of view, but instead use a simplified representation that allows TF-target dependencies to be used for interpretation. A network with 250 hidden nodes encodes TF-target associations. To derive interpretations, they independently alter the numerical value of network nodes corresponding to each transcription factor and observe the response at the output node that represent target genes.

Biologically-informed neural networks have important practical applications. They are especially becoming extremely useful in oncology [227], [228]), where they are used in studying plausible drug-tumor interactions and drug-drug interactions leading to new drug discovery and effective treatments recommendations. In addition, interpretation of various parts of the networks representing biological subsystems provides insights on the complex mechanisms by which these systems function. However, these approaches are currently limited by the problem of incomplete domain knowledge that makes it difficult to fully reverse-engineer complex biological mechanisms. Hence, the performance of these networks may not reach the levels of black-box models. Also, approaches that rely on altering individual factors to determine their effects (e.g., [226]) may not be able to capture possible synergetic interactions that are only observable when all influencing factors act together. Therefore, there is still a lot of room for improvement, not only from the perspective of network architectures, but also from the point of view of the very knowledge that informs their design.

Another important approach used to simultaneously improve interpretability and biological realism in deep neural networks relies on reducing the number of learned features to a manageable number made up of only those that carry biologically useful information. This concept is motivated by the fact that one of the main factors that accounts for the poor interpretability of deep learning models is their very large feature space—state-of-the-art CNNs typically have millions of features. Furthermore, as described in Section 5, CNNs are known to encode highly relevant psychological information that aligns with the brain's own representation of natural stimuli. However, it has been suspected that some of the CNN features may be redundant and only a subset may actually encode psychologically relevant information. Consequently, through similarity judgment and object categorization alignment, Jha et al. [215] propose to determine the lowest CNN feature dimensions needed to fully characterize human psychological representations and facilitate interpretability. In the case of similarity judgments, Jha et al. propose the SimDR architecture (Figure 13) which incorporates a small-width bottleneck layer to project the 4096 features of the last fully connected layer of a VGG-19 network [229] to a lower dimensional space that retains the ability to emulate human similarity judgments.

They observe that as few as six feature dimensions are sufficient to capture biological representations required to accurately predict human similarity judgments. They then rank the reduced dimensions in order of importance according to the extent to which each contributes to explaining the observed variance. Finally, the embeddings of each dimension can be inspected for interpretability and cognitive insights. For example, the study shows that for each dataset, the highest ranked dimension captures broad, semantically meaningful categorical information about the objects in the dataset—e.g., mammals vs non-mammals, animate vs inanimate objects, etc.—implying that human judgement relies heavily on these semantics.

Tarigopula et al. [230] notice that feature reprojection from high dimensional to low dimensional space, as proposed by Jha et al. [215], may cause information loss, leading to degraded performance and poorer interpretability. The authors hypothesize that the learned features in their raw form may already encode relevant information required for capturing human similarity judgments, and hence, do not need further transformation. They therefore propose to prune out redundant features, leaving only a small subset of relevant features. Their approach resulted in improved performance, implying that indeed, the last fully connected layers of CNNs already exclusively capture psychologically relevant information. Besides boosting explainability and biological alignment through their semantically meaningful psychological representations, the implications of these studies are also significant for the future of efficient neural network design. Rather than downscaling high dimensional models, new designs may employ lightweight architectures that directly learn only relevant features through appropriate optimization and training regimes. This could lead to significant savings in computational resources and training times. Additionally, the resulting models would run faster in deployment and offer real-time performance, which is currently lacking in state-of-the-art models.

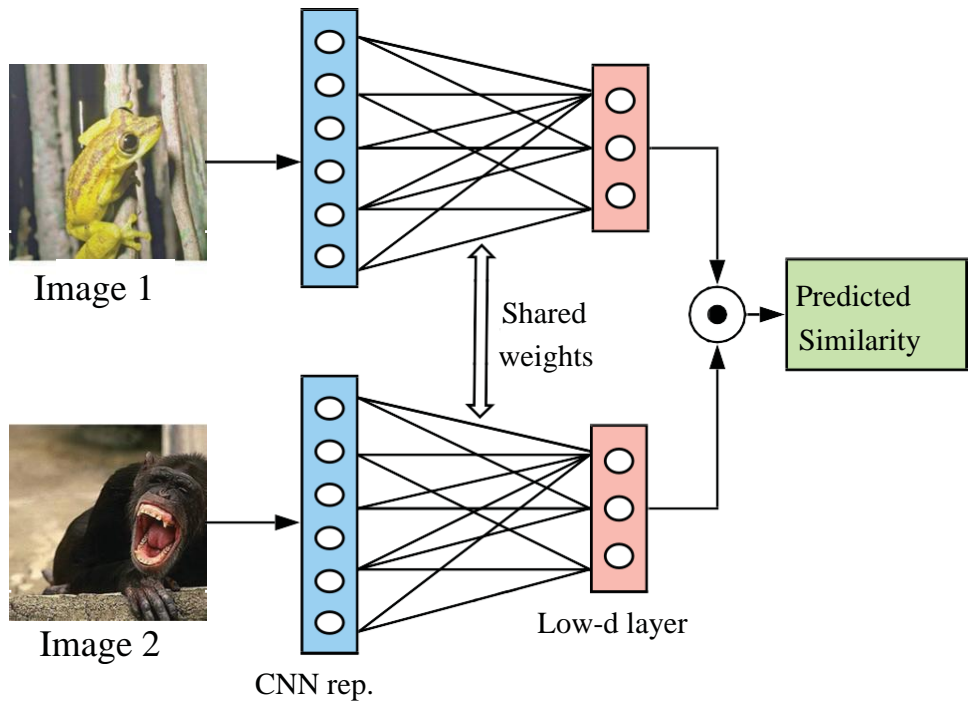

Figure 13. CNN architecture of SimDR [215]. The network takes a pair of images as input and a bottleneck layer projects the learned features to a low-dimensional space. The inner product of the resulting feature map is used to compute similarity scores. The resulting model is easier to interpret since it has a reduced feature set as it retains only biologically-relevant features.

### 4.3 Improving zero- and few-shot generalization with cognitive insights

#### 4.3.1 Cognitive architectures for zero- and few-shot generalization

Although, zero- and few-shot learning are not specifically implemented in cognitive architectures, this concept is implicit in the overall goal of autonomy and artificial general intelligence capabilities that some cognitive architectures attempt to achieve. Thus, an autonomous system must be able to recognize, understand and deal with situations even if it has little or no prior experience with the given situations. This capability is common in cognitive architectures designed for autonomous robots (e.g., iCub [202], IMPACT [204] and general-purpose systems designed to emulate human cognition (e.g., ACT-R [33], CLARION [231]). To meet high autonomy requirements, cognitive architectures typically employ a host of modules and subsystems that enable human-like learning, reasoning and adaptation to new environments. Some of the most important components needed to achieve this capability are: 1) a well-structured memory system used for storing knowledge and facilitating computations during reasoning and learning; 2) a multi-purpose learning system that integrates various learning paradigms to attain various competences; 3) a motivation-driven goal generation module to direct the activities of the cognitive agent; 4) an attention system that allows the agent to focus on the most important information for the given task.

The exact implementations of these components vary widely across different architectures. Some architectures like ACT-R and LIDA strive for high biological realism. In this way, the artificial agents do not only serve as proficient zero-shot learners and reasoners deployed to perform tasks autonomously in unknown environments, but they also serve as frameworks to study biological cognition. For instance, Juvina, and Taatgen [232] adapt ACT-R by adding a decaying characteristic to its memory retrieval mechanism and utilize the resulting system to model cognitive suppression that facilitates studies on negative priming, inhibition-of-return and the Stroop effect—a phenomenon that describes cognitive interference caused by a mismatch in the representation of a stimulus and its description or name (for example, when the name of a color is printed in a different color). Also, by fitting ACT-R model to model to functional Magnetic Resonance Imaging (fMRI) signals of the brain, Borst and Anderson [233] perform analysis to locate several neural processes, such as fact retrieval from memory and motor responses, retrieval of mathematical facts from memory, working memory updates, motor responses, and visual encoding. Therefore, besides overcoming zero-shot and few-shot generalization problems, development of biologically-informed cognitive architectures can help to create platforms necessary for conducting experimental studies leading to improvement in the understanding of the human brain. A better understanding of biological cognition would, in turn, enhance the advancement of artificial agents since these principles are needed to improve biological realism of AI systems.

### 4.4 Brain-inspired DNN for few-shot learning

Whereas zero-shot generalization significantly relies on reasoning in terms of the relations between observed and new categories, few-shot learning capacity is influenced greatly by the learning characteristics of a model, specifically, the ability to learn in a data-efficient manner. In biological systems, synaptic plasticity [235] allows the brain to form new connectivity patterns to enable it to encode new information while retaining existing relevant knowledge in memory. This facilitates data-efficient learning, allowing the brain to generalize well from only a few examples. However, the learning mechanism of gradient-based backpropagation algorithms commonly employed in contemporary machine learning is not compatible with the synaptic plasticity dynamics (see for example, Clopath et al. [236], which require error computations to be local to the neuron. Backpropagation relies on computing a global error vector by transporting gradients backwards through network layers. However, recent studies (e.g., [237]) have shown that this backward transport mode is not required to achieve competitive performance. Specifically, Baydin et al. propose to compute gradients exclusively in the forward mode using directional derivatives, and show that the forward gradient method can be faster in reaching a reference level of validation loss. It is also less computationally expensive and more biologically plausible compared with the strict requirement of backward transportation of errors that the backpropagation algorithm upholds.

On the other hand. predictive coding is a brain-inspired method that approximates the learning mechanism of the backpropagation algorithm [238], [239], [240]) using local error computations. Unlike gradient-based learning which performs weight updates of learnable parameters at all nodes using a single error computed at the final layer, predictive coding relies on minimizing the prediction errors associated with each neuron. A comparison of backpropagation- and predictive coding-based error computation is illustrated in Fig. 14.

Previous work [241] has demonstrated the effectiveness of associative memory in achieving efficient prediction by storing relatable memory vectors of learned features which can later be retrieved when an incomplete pattern similar to a stored instance is presented. Meanwhile, more recent stud-ies [242], [243] show that predictive coding actually implements an effective associative memory system. Expectedly, predictive coding-based networks have therefore demonstrated promising performance in data-efficient learning or few-shot generalization in deep neural networks.

More concretely, empirical results by Lee et al. [234] show the superiority of predictive coding over backpropagation across various few-shot generalization settings in DNNs. In the study, the predictive coding network consistently outperforms the model trained by backpropagation under a range of few-shot learning protocols. The study also shows promising incremental learning ability of the network trained with predictive coding, which is able to learn new information (in this case new object categories) without severely compromising the existing knowledge. On the other hand, a neural network that implements backpropagation on a new task setting may lead to the catastrophic forgetting problem [244], which causes the network to "forget" originally learned information when its weights are updated. Without any memory element to store previous information, catastrophic forgetting and poor few-shot generalization of standard backpropagation-based learning frameworks should not come as a surprise. Furthermore, local error computations in predictive coding networks can be carried out in parallel since they are not dependent on one another, unlike regular backpropagation-based errors which must progress in sequence from deeper layers to shallower ones. This independent weight update scheme can speed up the training process and allow optimization at the neuron level, potentially resulting in improved performance. Another way in which backpropagation may fair worse than predictive coding is the potential occurrence of accumulated rounding and other numerical computational errors caused by the sequential information processing.

Despite its strengths, predictive coding in deep learning is computationally demanding, and this raises serious difficulties when training large state-of-the-art neural network models. Yet, in the scheme of things, it can be argued that predictive coding is still attractive as a viable training method as it has also been shown to bring other benefits such as adversarial robustness [43] to deep learning systems. It should be noted that conventional adversarial training alone is computationally expensive, and combining adversarial training with few-shot learning goals would otherwise be very demanding to implement using conventional backpropagation-based training methods. Thus, in practical applications where artificial intelligent systems are required to exhibit good generalization and robustness properties, predictive coding can be leveraged to meet both objectives at the same time. This philosophy mimics biological systems better as they are not usually tailored for a single operational objective but instead can perform creditably under a wide range of situations.

Beyond training-level biologically-inspired enhancements, architectural-level modifications have also shown promising data efficiency, i.e., zero-shot and few-shot generalization capacity. For instance, Liu et al. [42] design the so-called Approximate Biological Neural Network (ABNN) to generate biologically plausible data used to fit two DNN variants – a Multi-Layer Perceptron (MLP) and a Long Short-Term Memory (LSTM) network. ABNN incorporates the Boundary Vector Cell model of the cortical neural circuit (structures credited with self-localization power in the brain) [245] and implements a host of other brain-inspired plausible features: different activation functions are employed to simulate non-uniform neuronal firing behaviors; dropout regularization models sparsity of neuron connections; injection of a small amount of Gaussian noise to the connection weights simulates plasticity. Other biologically plausible features include variable neural units, connections with feedback and lateral inhibitions. Empirical results confirm the effectiveness of the ABNN-trained models which achieve significant data efficiency. Although the degree to which each biological equivalent component benefits ABNN's data-efficient leaning ability has not been quantified, the fact that many of these modules may indeed play a major role in this behavior is well supported by current knowledge. For instance, cortical inhibition may preserve memory content [246] by preventing unwanted signals from corrupting it. This facilitates learning with few examples as some features learned can be captured in memory and would not be immediately lost when new knowledge is learned. The memory itself is facilitated by recurrent connectivity and enhanced by LSTM which captures long-term dependencies to enable the network to "remember" useful relationships needed for efficient learning and prediction. The study confirms, in line with existing knowledge on biological neural systems, that the ABNN-trained variant that combines recurrent connections with feedback and lateral inhibition and LSTM layer achieves better performance than the variants that employ MLP without these features.

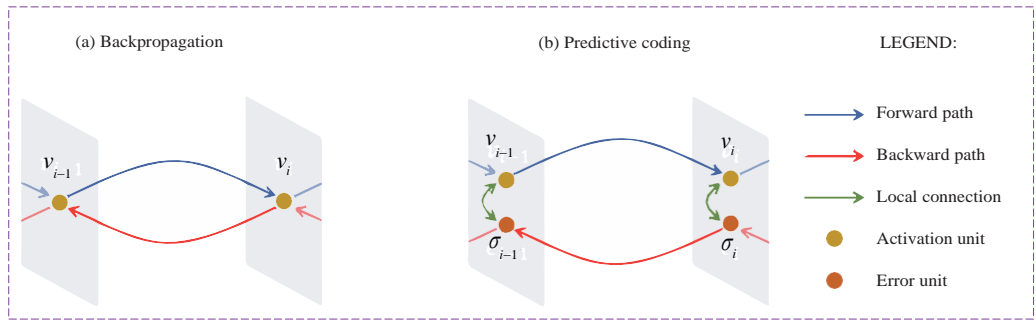

Figure 14. Difference between (a) backpropagation and (b) predictive coding [234]. Backpropagation computes a single error for the entire forward pass while predictive coding computes local errors for each activation unit.

Although the model employed in this study utilized oversimplified representations of various biological neural components, it nonetheless attains high data efficiency. Furthermore, the network variant that incorporates all the intricate features achieves the best performance, suggesting that zero-shot and few-shot generalizability in neural networks may be significantly enhanced by striving for more biological realism. This prospect is therefore exciting as there is still a wide scope for improvement in biological plausibility of this framework. For example, while the simulated neural inhibition mechanism proposed is exclusively localized between adjacent neurons, biological networks implement both local and global excitatory and inhibitory actions that help maintain stability at both levels.

## 4.5 Components of cognitive architectures commonly used to enhance adversarial robustness, interpretability, and zero-shot or few-shot generalization

To attain the competences needed for solving problems related to adversarial robustness, interpretability, and zero-shot or few-shot generalization, biologically-inspired cognitive architectures incorporate a wide range of features with suitable functionalities. In the literature, the most prominent among these features are: memory system, learning system, motivational system and attention mechanism. The features are usually implemented differently across different architectures. Their exact connections and relationships also vary widely from one architecture to another. Figure 15 shows, collectively, the main com- ponents required to achieve adversarial robustness, explainability and zero- or few-shot learning. Table 2 then presents the specific capabilities each of these components may enhance. We briefly describe the important features of these components next.

**Memory system:** To ensure adversarial robustness and zero- or few- shot generalization ability, a well-developed memory system embodying various parts that facilitate a variety of functions is required to store and maintain knowledge for effective decision-making: To enable easy and speedy access, working memory (WM)—a form of short term memory (STM)—temporarily holds the information that is currently being processed. The arrangement enables fast decision-making. After processing, the information is transferred to long-term memory (LTM) for a lasting storage as acquired knowledge. Long-term memory may store information as symbolic instructions or rules, with relevant metadata that describes context, relations or any other important background information that may be useful in decision making. In addition to symbolic elements, neural network weights may also be stored in long- term memory.

The functional components of LTM are 1) procedural memory, used for storing implicit knowledge learned from spontaneous behaviors like swallowing; 2) declarative memory, which is made up of semantic memory used for storing factual information about entities and their relationships; and 3) episodic memory for storing experiential knowledge acquired over time. ACT-R represents working memory as a collection of separate memory buffers to support parallel processing by individual modules. Temporary results of each model is stored in the module-specific WM buffer and a symbolic module coordinates the activities of these units. Some cognitive architectures—as in the case of ACT-R— usually store episodic memory information in the form of "stimulus, decision, outcome", and use it to guide behavior.

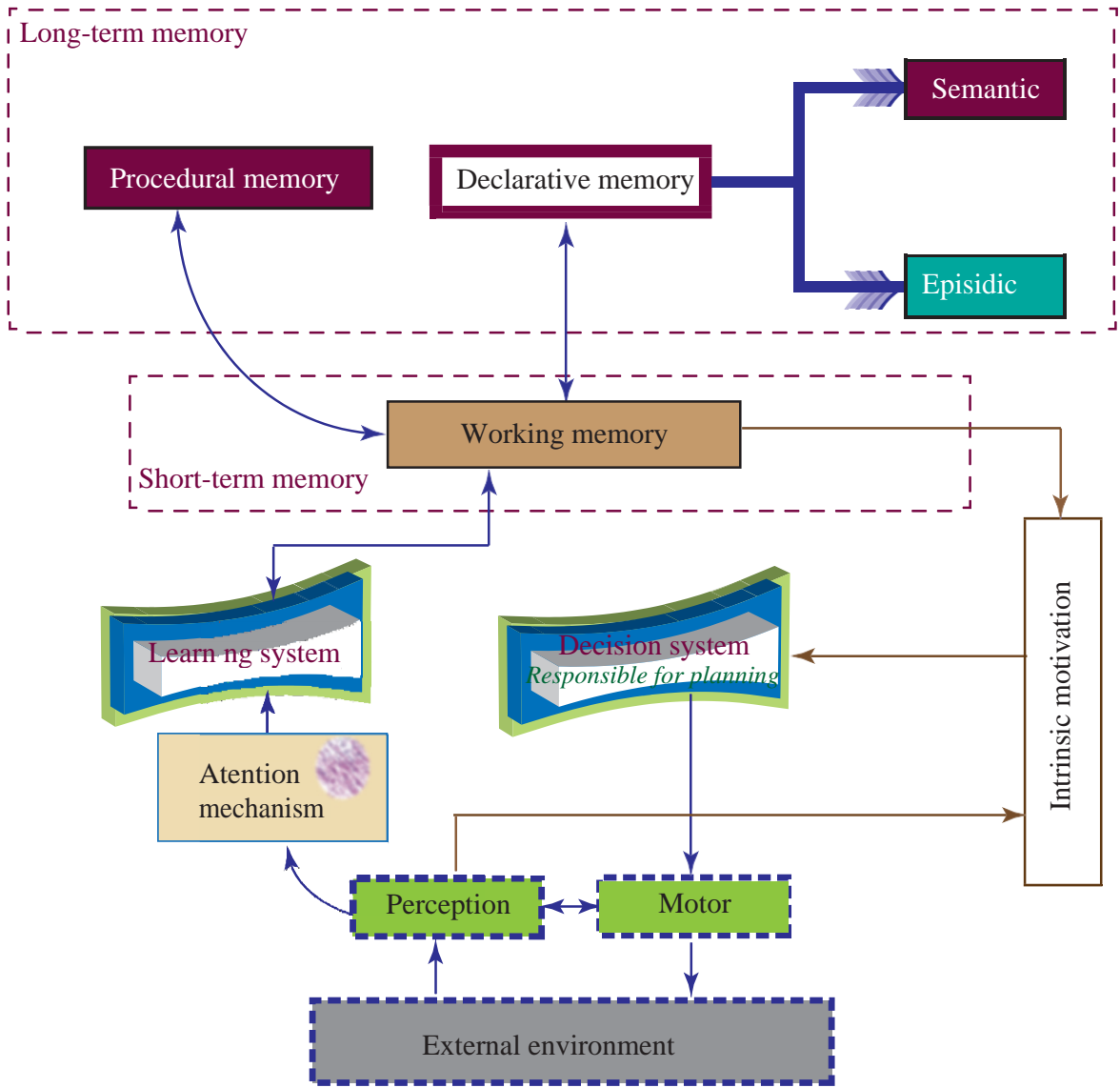

Figure 15. Main components of a biologically-inspired cognitive architecture.

**Learning system:** Zero-shot and few shot learning capabilities, as well as high robustness are facilitated by a corresponding memory system that allows storage and retrieval of different forms of knowledge acquired through a powerful learning process. The learning should progress incrementally, without new information overriding existing knowledge, but instead used in enhancing it. Online learning may result in the update of components like neural network weights, symbolic instructions or rules, or associated metadata. An entirely new experience may involve the formation of new symbol structures. In the process, corresponding changes are made to procedural and declarative long- term memory contents.

The ability to learn online is imperative, as this feature enables the agent to adapt its behavior by learning from mistakes and good decisions alike. For effective zero-shot and few shot learning, multiple learning methods may be combined to enable the acquisition of different kinds of skills.

For example, in ACT-R and IMPACT, procedural learning is accomplished by composing new rules from example observations. This allows new skills and knowledge to be formed through experience. At the same time, these architectures implement associative learning through the action-reward paradigm, also known as reinforcement learning. Reinforcement learning facilitates adaptive behavior by enabling the agent to learn useful associations between actions and their outcomes.

**Motivational system:** Effective learning also requires a well-developed motivational system that enables intelligent agents to autonomously generate new goals and determine or select appropriate actions to achieve these goals in the presence of environmental constraints. A goal is usually a new state the agent would like to be in, and the motivation provides the reason for striving to be in that state. For example, IMPACT is a robotic platform designed for exploration. It can autonomously generate goals driven by motivations such curiosity and novelty that inspire it to discover new states by exploring its environment. In cognitive architec-tures, goals are usually generated to maximize the agent's overall pay-off (which is inherent in the motivation) once attained. These capabilities enable autonomous robots to learn new skills and acquire new competences needed for operating and achieving desired success in their environments.

Table 2
Main features of cognitive architecture and the objectives they can help to meet. Meaning of symbols: √ means the feature helps in attaining the particular objective; − means negative impact; and × means the objective is unaffected by the feature.

| Feature | Objective | | | Highlights |
| --- | --- | --- | --- | --- |
| | XAI | Adv. | ZSL/FSL | |
| Multisensory perception | × | √* | × | *Multisensory perception helps to conduct sanity checks in order to detect when any modality is inconsistent with the others as a result of adversarial attack. |
| Symbolic processing | √* | | | *Symbolic processing provides interpretable rules and are the main source of interpretability in most cognitive architectures. |
| Online procedural learning | −* | √ | √ | Online procedural learning involves learning from examples or rules encountered through experience. Additionally, reinforcement learning can be leveraged to provide weights over choices so as to improve the learning outcome.<br>Decisions made by the learning module preclude explanation, unless the knowledge acquired through learning is used to consolidate symbolic rules which in turn provide explanations. |
| Online associative learning | −* | √ | √ | Online associative learning (usually formulated as reinforcement learning problem) involves the gradual learning of skills through repeated interactions with the environment.<br>Out-of-the-box, associative learning does not facilitate exlainability. |
| Working memory (WM) | √ | √ | √ | Short-term or working memory store temporary information that is essential for the requisite computations during decision-making. |
| Long term memory (LTM) | √ | √ | √ | LTM stores the symbolic structures that encode acquired knowledge (e.g., facts) as well as the associated metadata that provide guidelines on the use of the knowledge. Both short-term and long-term memory systems are essential for meeting the capabilities required for explainability, adversarial robustness and zero- or few-shot learning. |
| Intrinsic motivation | √ | √ | √ | Effective motivational systems facilitate adaptability and robustness as they can drivecognitive systems to generate new set of goals in response to changing environments, and select appropriate actions to achieve these intrinsic goals. Hence, they offer adversarial robustness and ZSL/FSL capabilities. |
| Attention | √ | × | × | Attention maps can provide explanations by highlighting the most salient regions of the percept that influence a given decision. This is common in visual perception (e.g., [247] and [248]. |

**Attention mechanism:** To manage computational and memory overload, some cognitive architecture frameworks like CLARION and iCub implement attention mechanisms that enable them to focus on the most important information. For instance, in these cognitive architectures, data is received from multiple sensor modules but only the most relevant information to the task at hand is immediately processed. This prevents distractions and allows the agent to attend to the most important events quickly. Summary: All the components and features of cognitive architectures described in this section and in the preceding sections work in combination to meet the operational needs of intelligent agents that utilize them. To function satisfactorily in the open-world, these agents must satisfy key performance objectives. First, they should be able to make decisions in a zero- or few shot manner, i.e., in situations they have little prior experience in. This goal is usually achieved by leveraging the capabilities of learning and reasoning modules in tandem. Additionally, these agents must exhibit robustness to unusual and adversarial sensory inputs. Robustness is aided by access to multiple sensory cues. Finally, for applications in collaborative robotics and simulations of biological cognition, agents must be able to explain their decision logic. Usually, symbolic components provide this interpretability, although learning components based on neural networks that leverage explainable artificial intelligence techniques can independently provide explanations on their predictions over sensory input.

Presently, most cognitive architectures do not explicitly specify adversarial robustness, interpretability and zero- or few shot generalization objectives. Instead, their motivational systems generate goals that maximize utility of overall task performance, which implicitly could include any combination of these objectives. The lack of explicit requirement on these objectives makes it difficult to compare models on the effectiveness of their realization. In fact, it is sometimes hard to access whether a given model achieves some of these objectives. Consequently, some of the descriptions of cognitive architectures in the context of adversarial robustness, interpretability and zero- or few-shot generalization are based on direct relationships between their actual capabilities and the capabilities needed to achieve these specific objectives. The components of cognitive architectures discussed in this section are critical in attaining these capabilities.

## 5 BRAIN DECODING USING PRE-TRAINED DEEP LEARNING MODELS

Brain decoding allows the content of the brain— in this case the brain's response to visual stimuli—to be decoded by transforming it into semantics-rich images, facilitating a deeper understanding of how the brain encodes visual information. Eventually, it can also provide clues on *what is on the mind*, and help artificial agents to understand their human counterparts in human–computer interactions (HCI). And in conjunction with explainable AI techniques that enable humans to understand the decisions of machine learning systems, mutual understanding between humans and AI systems in collaborative tasks can be attained.

Implicit knowledge captured by pre-trained foundation models has proven useful in brain decoding. Specifically, CLIP [163] has shown the most exciting prospect in reconstructing natural scenes from functional Magnetic Resonance Imaging (fMRI) signals of the brain. For instance, BrainCLIP [249] achieves impressive results by fine-tuning CLIP on fMRI signals with additional image and text data provided as prompts. The zero- and few-shot generalization property of CLIP which has been confirmed by several studies (e.g., [250], [251]) is critical in the open-world neural decoding task. Previously, research aimed at decoding visual stimuli from brain activity recordings mainly concentrated on unimodal approaches that rely on either natural language descriptions of viewed images (e.g., [252], [253],

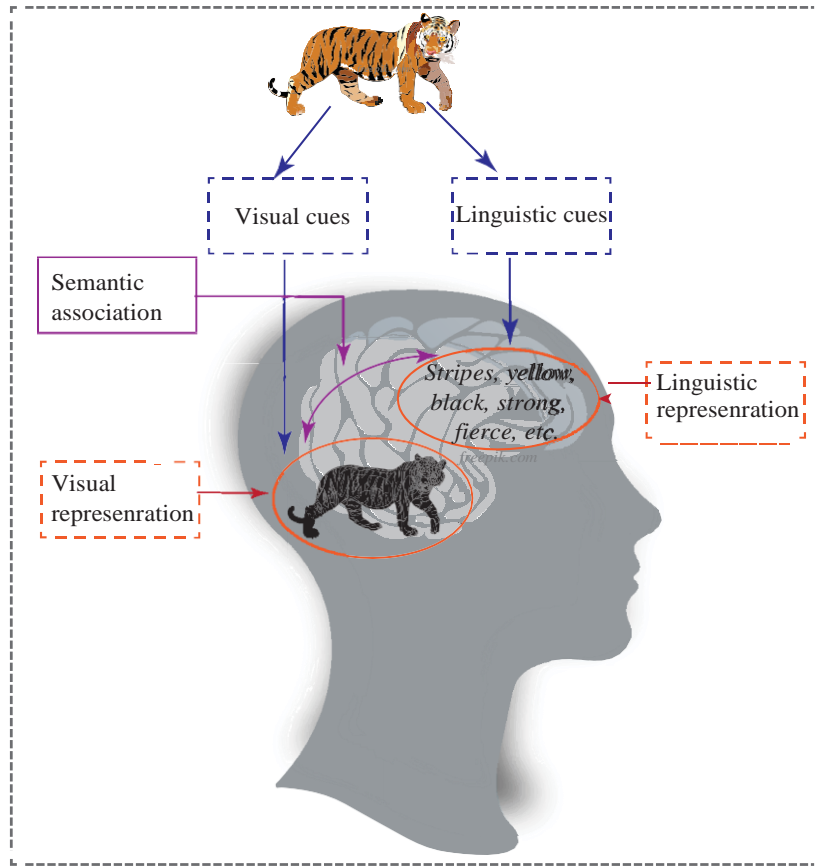

Figure 16. Dual coding account of scene representation: In addition to the visual impression, intuitive linguistic attributes associated with the objects present in the scene are also registered in the brain.

[254]) or visual semantic information (e.g., [255]). These methods learn representations that map raw fMRI signals to textual representations or natural images. Although these approaches achieve satisfactory results in simpler tasks like visual stimuli classification –that is, categorizing the image that produces the given fMRI response – reconstruction of whole scenes with high semantic fidelity remains a major challenge. More recently, a number of studies (e.g., [43], [249]) have established that multimodal vision-language frameworks perform far better in this task than unimodal approaches due to a better representational ability of the former.

This assessment is consistent with the conclusions of the dual coding theory ( [256], [257]), which explains that the human brain efficiently encodes visual information on natural scenes using a combination of verbal and visual representations. According to this view, as shown in Figure 16, when an observer sees a tiger for example, several associated verbal concepts (such as yellow, stripes, strong, fierce) that have been established through experience and commonsense knowledge about the object are automatically registered in the brain. The corresponding dual image-linguistic representation encodes a wealth of background information with semantic relations to the content of the scene. This representation is far richer than what either modality could possibly capture alone. Unsurprisingly, therefore, the methods that adopt multimodal frameworks with the ability to simultaneously represent both visual and verbal concepts have yielded better results.

Another factor that accounts for the performance gap between regular and pre-trained models is the lack of sufficient quantities of training data. Producing sufficient quantities of fMRI responses to external stimuli is an extremely challenging activity. Hence, extensive datasets of paired stimuli-responses for training brain decoding networks is a difficult task. Eventually, models trained on the limited data would need to handle unseen samples (or those with only a few training examples) when deployed in-the-wild. Therefore, the neural decoding task is essentially a zero- or few-shot generalization problem. Fortunately, aided by their zero- or few-shot generalization ability, pre-trained foundation models, particularly CLIP-based brain decoding frameworks, have demonstrated strong neural decoding performance in the zero-shot setting.

The emergence of several state-of-the-art models like BrainCLIP [249], MindEye [258], Mind-Reader [259] and Brain-diffuser [260] is a strong evidence to this fact. These models are first trained to map raw fMRI data to the aligned image and text embedding space of the CLIP model through contrastive learning. The resulting representation is then used to reconstruct the original scene that generated the brain response captured in the fMRI signals. For scene reconstruction, image generation are required to transform the aligned CLIP embeddings to natural images. The commonest image generation methods employed are Generative Adversarial Networks (GANs), Variational Autoencoders (VAEs), and Diffusion Models (DMs).

Figure 17 shows sample results of visual scenes gen-erated from input fMRI signals using Mind-Reader [259] The results show that while the appearances of objects reproduced by the model do not exactly match the original images, the overall high-level semantic content of the scene is preserved. This impressive performance on decoding brain signals to reveal visual information in the form of photorealistic images has the potential of helping to accelerate studies on how the brain represents visual information.

# 6  DISCUSSIONS

Conventional deep learning models are designed for specific applications under a set of (sometimes unrealistic) assumptions. Humans on the other hand, can learn from a variety of contexts and connect the acquired knowledge in powerful ways to solve even unfamiliar problems. Yet, owing to their remarkable performance in designated applications, current deep learning methods typically lack compelling reasons to incorporate additional design or training innovations that leverage human understanding beyond the training data, as long as they operate in appropriately-constrained domains. This explains why most of the current state-of-the-art models are the data-driven type. However, motivated by the need to solve practical problems concerning general and adversarial robustness, model explainability and zero- or few-shot generalization, research on augmenting deep learning models with prior domain knowledge and brain-inspired techniques is generating a lot of interest. In this review, we extensively explore these methods of alleviating the aforementioned challenges in practical applications. We also highlight ways in which knowledge-informed and brain-inspired artificial intelligence techniques are used to improve human under- standing of biological cognition.

1)  **Prior knowledge-informed deep learning methods**

a)  *General Overview:* In all three problems considered in this paper – adversarial robustness, interpretability, and zero-shot or few-shot generalization – methods utilizing logical rules are overwhelmingly popular and effective among knowledge- informed approaches. A few works incorporate knowledge graphs but most of these additionally leverage logical reasoning to make their decisions. This is unsurprising, since in most practical situations the underlying symbolic logic can capture guiding rules to help deep learning models discover and correct illogical decisions. This feature is especially useful in combating adversarial attacks. Also, in the realm of explainable AI, knowledge graph-based approaches are the predominant techniques. This is partly due to the fact that the application domains that mostly require explainability (e.g., medical diagnosis, recommender systems, and visual question answering) tend to have much of their training dataset already represented as knowledge graphs and ontologies. That said, knowledge graphs are, in fact, easy to extract explanations from since they capture contextual relationships among entities. Again, logical rules are commonly used to augment knowledge graphs so as to provide valid explanations. They are also commonly used as standalone solutions when the data is not already represented as knowledge graphs. Yet again, logical reasoning and knowledge graphs are effective in zero-shot learning, where their strengths lie in the fact that new concepts can be recognized if knowledge about previously learned concepts, together with their relationships to new or unseen concepts is provided. Meanwhile, knowledge graphs elegantly represent these relationships while logical reasoning plays an important role in connecting various facts among observed and unseen concepts.

Knowledge in probabilistic representation, physics-informed deep learning and explicit mathematical relations are the least utilized knowledge forms in solving problems relating to adversarial defenses, explainability, and zero-shot learning. Mathematical and physics-based laws usually improve accuracy and out-of-distribution robustness. The low utility of probabilistic reasoning in these applications can be explained by the fact that this form of knowledge is not able to capture requisite back- ground context needed to disambiguate subtle situations. Whereas physics-based representations and explicit mathematical relations can encode theoretically grounded information, their inability to capture semantically-faithful relations about the environment accounts for their relatively low popularity.

b)  *Adversarial robustness:* The success of future adversarial defense strategies may lie in the ability to conceal the internal mechanisms of these techniques from potential adversaries. Even with knowledge informed defenses, if the rationale behind the defense is known, effective counter-attacks are possible. The attackers can exploit knowledge of the underlying mechanism to create adversarial decisions consistent with the knowledge used by the defense. Despite the risk, efforts aimed at minimizing this possibility are practically non-existent. We expect future works to focus more on tackling this problem. A possible research direction could aim to equip knowledge-informed defenses with "adversarial smokescreens" that will conceal their true inner workings and instead trick potential adversaries into believing they employ a different logic.

c)  *Explainability:* Various forms of knowledge representation are employed to derive explanations for the predictions of deep learning models. Some are concerned with enhancing the interpretability of conventional frameworks. For example, human-in- the-loop methods are often utilized to refine the heatmaps generated by gradient-based methods. Other knowledge- informed methods compose explanations directly from the models they explain. More significantly, knowledge-informed methods represent explanations in different forms such as natural language statements, logic expressions and concept values. It is therefore challenging to compare explainability results across different frameworks. Objective methods of quantifying the quality of explanations across different methods through unified performance metrics would be helpful in guiding practitioners to select appropriate methods for their applications. Also, to enhance some human-in-the-loop explainability methods, future work may need to ensure mutual understanding between humans and deep learning systems. Current work shows clear effectiveness of incorporating common sense or expert knowledge into deep learning models through direct human participation in the training or inference stages. These human-in- the-loop approaches are particularly useful for improving explainability. Here, human knowledge is usually employed to correct any wrong explanations given by a model, and since in some approaches the accuracy of the model's decision is contingent on the corresponding explanation being correct, this technique simultaneously improves explainability prediction ac- curacy. In human-in-the-loop frameworks, humans may interact with machine learning models through symbolic logic, natural language, or image annotations. Correcting model explanations through simple interfaces like logic rules is precise and unambiguous. However, it is currently difficult for the human participant to communicate exact intentions through saliency annotation because the annotation style only captures an imprecise representation of the user's intended input. It would be helpful for machine learning methods to be able to recognize the exact region intended by the user. One possible way of achieving this could be through annotation-specific regularizations during training. Also, in cases involving long training cycles, it is extremely laborious for humans to provide the needed debugging. Fortunately, a new line of works that seeks to engineer special artificial agents – called cognitive clones [261], [262]– capable of mimicking a individual's behavior and actions. These agents would be able to replace humans in the heatmap annotation process, thereby alleviating the burden in training this class of human-in-the-loop interpretability models.

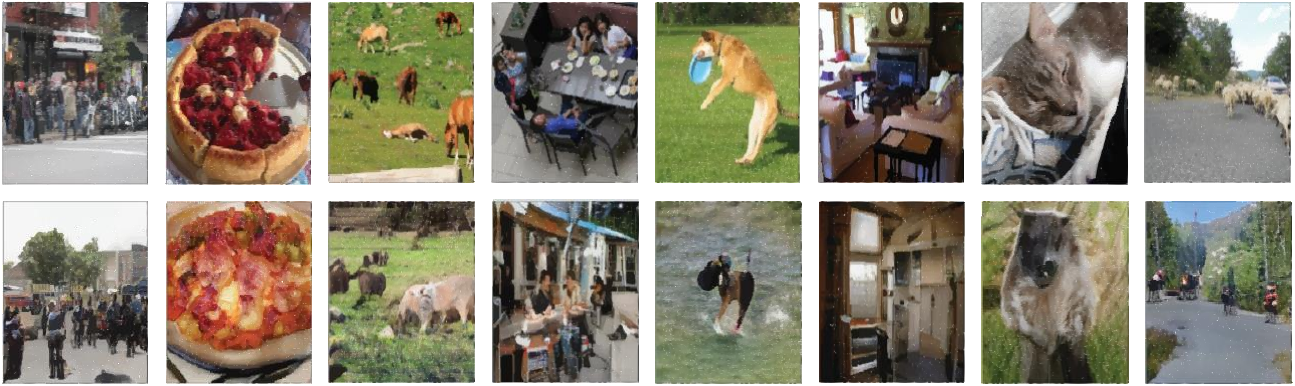

Figure 17. Sample results of scene reconstruction from fMRI signals. The op row shows the original image that generated the given fMRI response while the bottom row shows the corresponding reconstructed scene using Mind-Readern.

d) *Zero-shot and few-shot generalizations:* Many knowledge- informed approaches have achieved impressive results in generalizing to unseen instances. Implicit knowledge encoded by pre- trained foundation models has particularly demonstrated the ability to reach impressive zero-shot and few-shot generalization performance across different tasks, especially in question answering and visual question answering. This high generalization ability raises a major concern that borders on ethical and privacy matters. Specifically, as a result of pre-training on unrestricted amounts of data, these models acquire substantial world knowledge that can often be leveraged by end-users to generate privacy-infringing or ethically questionable content in ways that are unanticipated by their developers. Therefore, appropriate measures to ensure their proper use will be necessary.

### 2) Brain-inspired deep learning and cognitive architectures.

a) *Cognitive architectures:* In most cases, the effectiveness of cognitive architectures in handling the aforementioned problems lies in

the fact that they strive for a more general human-like intelligence characteristic. Hybrid cognitive architectures are among the most ambitious models that attempt to achieve artificial general intelligence. They incorporate many features of the brain and adopt a similar problem-solving paradigm. Specifically, in line with the dual process theory of human cognition, they employ two separate systems for this purpose – a learning-based system, named system 1, for fast decision making; and a reasoning-based system, named system 2, for slow and analytically grounded problem solving.

However, whereas in biological cognition each of these systems is adapted to play its intended evolutionary role without any conscious effort, the mechanisms that regulate this function in brain-inspired cognitive architectures are far from optimal. Some recent implementations deploy a third system to play a supervisory role, in some cases (e.g., [263]) this involves selecting the solver between systems 1 and 2 without considering the suitability for the problem. This approach may lead to sub- optimal solutions being produced. Instead, an AI system should understand the scope of its own knowledge in advance, and activate the appropriate system based on the problem at hand, or immediately engage system 1 if promptness is a critical requirement. If the mechanisms that control the functions of systems 1 and 2 are methodically evaluated and improved along the lines of biological plausibility, performance could be enhanced and better simulation models developed to gain further insight into how humans carry out these functions. This may even lead to outcomes that allow humans to overcome current deficiencies such as the law of reversed effort [264] and tendency to choke under pressure [265], [266]) that currently characterize human decision making when stakes are high. These deficiencies are associated with the fact that the reasoning system takes over control of decision making when in fact the task could be better handled by the spontaneous learning system.

b) *Brain-inspired neural networks:* Brain-inspired neural net- works aim to leverage the structure and information processing mechanisms of the brain to improve deep learning models. Some approaches make architectural enhancements to neural networks by incorporating modules that mimic various components of the brain. Other methods achieve biological plausibility by implementing training-level alignment through multi-task learning. Still, other techniques rely on similarity judgement optimizations to align the representations of neural networks and humans. In- formation processing techniques like predictive coding have been realized. The resulting models obtained from all these biologically-informed modifications show high performance in difficult settings compared with corresponding frameworks that do not employ these techniques. In particular, brain-inspired neural networks excel in adversarial robustness and zero- or few-shot generalization performance. Meanwhile, experiments based on similarity judgements have shown that although representational alignment – which characterizes the level of agreement between the representations of humans and DNN models – enhances adversarial robustness and few-shot generalization machine learning, the improvements are not absolute. In particular, models with low degree of alignment perform better in adversarial robustness and few-shot generalization tests than those with moderate level of alignment. How- ever, highly aligned models outperform both lowly and moderately aligned models. Although the exact reasons for this behavior are not yet established, we posit that this may be caused by the fact that moderate alignment may be enough to disrupt the original representational structure of the neural network but insufficient to instill human representational attributes needed to ensure human-like performance. Studies also reveal that alignment outcomes are dataset dependent. The implication of these findings is that developers of deep learning models seeking representational alignment must empirically determine the appropriate levels of alignment needed to achieve the required performance improvements. One of the main challenges of brain-inspired neural network design is the complexity of the brain. This makes it difficult to mimic most components accurately. Furthermore, human knowledge of the brain is still incomplete. Therefore, neural network designers can only achieve a limited degree of biological realism. Hence, neural networks still vastly underperform their human counterparts in adversarial robustness and zero-or few- shot generalization problems. However, the ability of biologically plausible deep neural networks to simulate behavioral phenomena associated with human task performance has opened up the opportunity to better understand the cognitive process. In contrast with ideal observer analysis which relies on rigid mathematical modeling to test various aspects of cognition, brain-inspired neural networks are flexible and, though training, are able to adapt their properties and representations to mirror the behavioral characteristics of human subjects performing a given task. Although, the exact extent to which deep learning models can improve their performance by leveraging knowledge of the brain is not yet known, results from current research on brain-inspired neural networks suggest that there is a wide scope for improvement in this direction.

3) *Zero- and few-shot brain decoding using pre-trained models* Importantly, for the neuroscience community, implicit knowledge in pre-trained models has shown a high potential for enabling the decoding of visual stimuli encoded by the brain. In particular, the high zero- and few-shot generalization ability of CLIP is leveraged to efficiently utilize low volumes of functional magnetic resonance imaging data — since this data is not available in large quantities owing to the difficulty in recording samples — to perform open-world brain decoding and scene reconstruction. Although current approaches achieve impressive results in terms of their ability to recover global semantic information in the reconstructed scene, object-level details for complex scenes. One reason that could account for this deficiency is the fact that these brain-decoding neural networks currently seek to recover the entire stimuli signature from only the visual cortex. However, while the visual cortex plays a fundamental role in visual stimulus encoding, other brain regions may also handle useful aspects of visual information needed recover the original scene in its entirety. Therefore, future work could incorporate relevant signals from additional parts of the nervous system to improve decoding accuracy.

# 7 CONCLUSION

Artificial intelligence systems have already reached impressive performance levels in general application settings. However, in special situations, these models significantly fail to achieve satisfactory results. We consider adversarial robustness, explainability and zero-shot or few-shot generalization, which are among the most important problems faced by deep learning systems today. In this paper, we review the approaches of tackling these problems. Specifically, we focus exclusive on the methods that leverage prior knowledge and those that are motivated by the brain's mechanisms of operation. Both lines of approaches have achieved significant success in addressing these problems. Our work also covers studies that aim to leverage the capabilities of artificial intelligence in advancing cognitive science knowledge. Again, there are important accomplishments in this direction. Despite this progress, there is still a large gap between the performances of artificial intelligence models and humans under challenging situations. Also, a lot more is yet to be discovered in cognitive science. Overall, given the current directions of research and the pace of progress, the future prospects of both areas are bright.

## ACKNOWLEDGMENTS

The authors would like to thank...